\documentclass[journal]{IEEEtran}

\usepackage{amsmath}
\usepackage{amssymb}
\usepackage{pifont}
\usepackage{subfigure}
\usepackage{graphicx}
\usepackage{url}
\usepackage{cite}
\usepackage{multirow}
\usepackage{makecell}
\usepackage{booktabs}
\usepackage{bm}
\usepackage{graphics} 
\usepackage{epsfig}
\usepackage{balance}
\usepackage[ruled,linesnumbered]{algorithm2e}
\usepackage{longtable}
\usepackage{colortbl}
\usepackage{float}
\usepackage{color,xcolor}
\definecolor{light-gray}{gray}{0.9}
\definecolor{mycolor}{RGB}{0,0,0}
\usepackage[colorlinks,linkcolor=blue,citecolor=blue,urlcolor=blue]{hyperref}

\begin{document}
\title{ID-Guard: A Universal Framework for Combating Facial Manipulation via Breaking Identification}

\author{Zuomin Qu, Wei Lu, \IEEEmembership{Member,~IEEE}, Xiangyang Luo, \IEEEmembership{Member,~IEEE},  Qian Wang, \IEEEmembership{Fellow,~IEEE}, Xiaochun Cao \IEEEmembership{Senior Member,~IEEE}
\thanks{This work is supported by the National Natural Science Foundation of China (No. 62441237 and No. U2001202). (Corresponding author: Wei Lu)} 
\thanks{Zuomin Qu and Wei Lu are with the School of Computer Science and Engineering, Ministry of Education Key Laboratory of Information Technology, Guangdong Province Key Laboratory of Information Security Technology, Sun Yat-sen University, Guangzhou 510006, China (e-mail: quzm@mail2.sysu.edu.cn; luwei3@mail.sysu.edu.cn).} 
\thanks{Xiangyang Luo is with the State Key Laboratory of Mathematical Engineering and Advanced Computing,  Zhengzhou 450002, China (e-mail: luoxy$\_$ieu@sina.com). }
\thanks{Qian Wang is with the School of Cyber Science and Engineering, Wuhan University, Wuhan 430072, China (e-mail: qianwang@whu.edu.cn). }
\thanks{Xiaochun Cao is with the School of Cyber Science and Technology, Shenzhen Campus, Sun Yat-sen University, Shenzhen 518107, China (e-mail: caoxiaochun@mail.sysu.edu.cn).}
    }

\maketitle
 
\begin{abstract}
    The misuse of deep learning-based facial manipulation poses a serious threat to civil rights. To prevent such fraud at its source, proactive defense methods have been proposed that embed invisible adversarial perturbations into images, disrupting the manipulation process and rendering the forged output unconvincing to observers. However, non-targeted disruption of the output may leave identifiable facial features intact, potentially leading to the stigmatization of individuals. In this work, we propose a universal framework for combating facial manipulation, termed ID-Guard. The framework employs a single forward pass of an encoder–decoder network to generate cross-model transferable adversarial perturbations. We introduce a novel Identity Destruction Module (IDM) to suppress identifiable features in manipulated faces. The perturbation generation is optimized by formulating the disruption of various manipulation types as a multi-task learning problem, with a dynamic weighting strategy designed to enhance cross-model performance. Experimental results show that ID-Guard effectively defends against diverse facial manipulation models while degrading identifiable regions in manipulated images. It also enables disrupted images to evade facial inpainting and facial recognition systems. Moreover, ID-Guard can be seamlessly integrated as a plug-and-play component into other tasks, such as adversarial training. The source code is publicly available at \url{https://github.com/ZOMIN28/ID-Guard}.
\end{abstract}


\section{Introduction}
\IEEEPARstart{T}{he} spread of false information within communities has long been a major concern, posing significant threats to civil rights and social security. The rapid advancement and widespread adoption of generative deep neural networks (DNNs) have further exacerbated this problem, with facial manipulation emerging as a prominent example. This technology enables the end-to-end manipulation of facial attributes or identities in images and videos. Malicious actors, for instance, exploit forged images to fabricate and disseminate misleading news~\cite{shu2020fakenewsnet,nan2021mdfend} or to commit online fraud~\cite{davey2023deepfake}. Although retraining these models is still challenging due to substantial computational demands and technical complexity, pre-trained models are readily accessible on open-source platforms such as GitHub~\footnote{https://github.com}, Hugging Face~\footnote{https://huggingface.com}, and TensorFlow Hub~\footnote{https://www.tensorflow.org}, allowing users to easily perform forgeries~\cite{aneja2022tafim}. This accessibility dramatically lowers the barrier to producing fake content, thereby accelerating the spread of misinformation on social media. There is therefore an urgent need to develop effective and proactive defense mechanisms.

    \begin{figure}[]
          \centering
          \includegraphics[width=\linewidth]{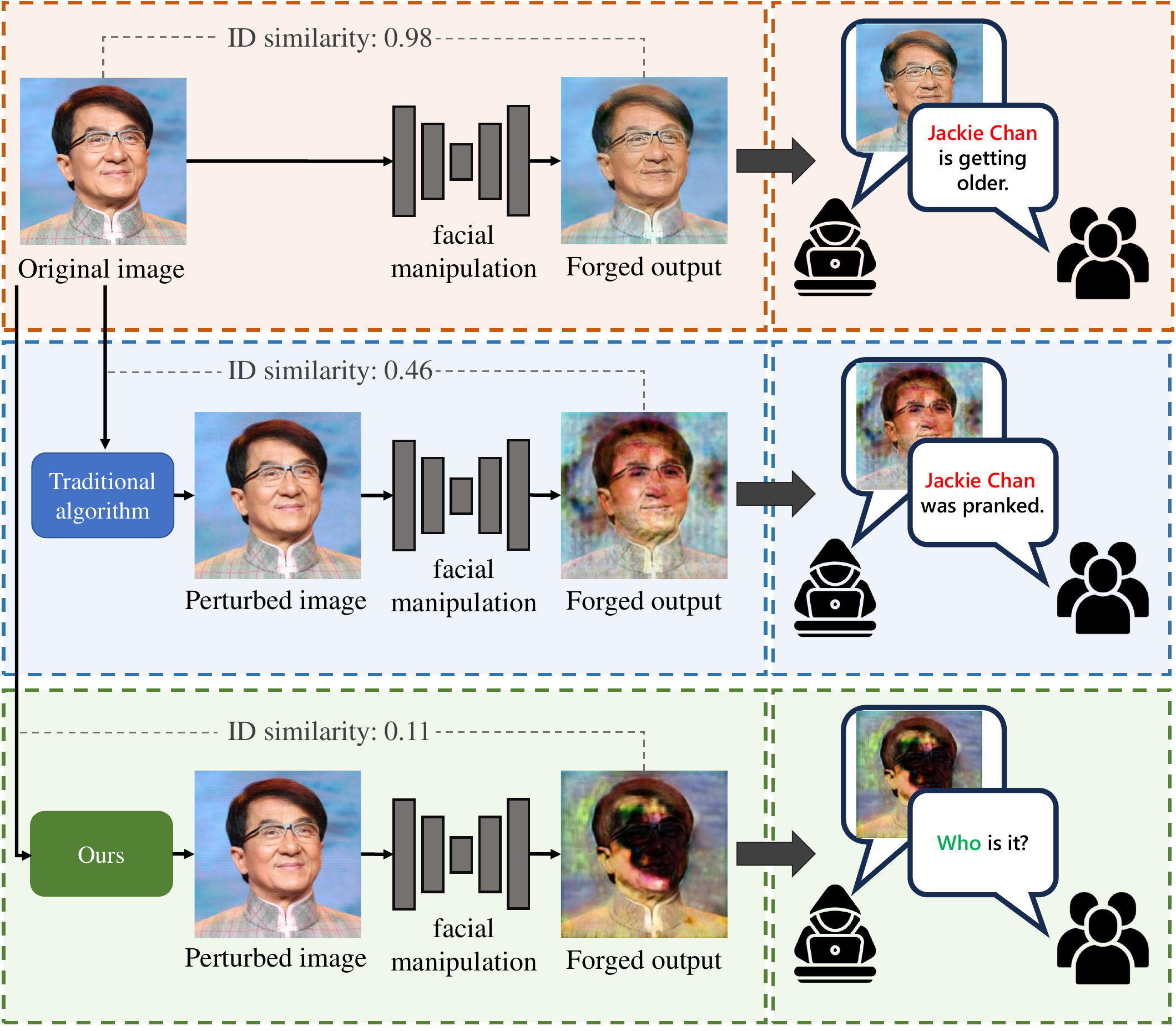}
          \caption{Illustration of the impact of malicious propagation of facial manipulation samples. Fakes will lead to rumors spreading, and insufficient distortion of faces by traditional defense methods will cause face stigmatization. Our method disrupts the observer's identification of the identity in the sample and thus adequately protects the individual's rights.}
          \label{fig:intro}
    \end{figure}

In response to these threats, considerable research has recently focused on developing proactive defense mechanisms against facial manipulation. Unlike passive detection methods \cite{yin2023dynamic,liao2023famm,zhang2022unsupervised,qiao2024fully,zhu2023face}, proactive defense methods \cite{ruiz2020disrupting,yeh2020disrupting,huang2022cmua,aneja2022tafim,guan2022defending,dong2023restricted,zhu2023information,zhai2023defending,tang2023feature,qu2024df,guan2024adversarial} aim to counter fraudulent activities at their source.
However, these approaches face several critical limitations: 1) they often fail to fully obscure personally identifiable features, allowing identity-relevant information to persist, since the distortions produced by unconstrained adversarial perturbations tend to appear in indeterminate locations rather than effectively covering the entire face; 2) they introduce random and unstructured artifacts such as shadows, background distortions, and deformed facial regions, which produce unnatural and visually unappealing results. Consequently, if persistent malicious users upload these disrupted images to social networks and they are widely disseminated, concerns about facial stigmatization arise, posing ethical and reputational risks to the individuals depicted~\cite{zhai2023defending}. This issue is particularly pronounced for public figures, whose identities remain recognizable even after defense mechanisms are applied. For example, as shown in Fig.~\ref{fig:intro}, when traditional proactive defense is applied to a photo of the famous actor Jackie Chan, his identity remains discernible despite the visual distortions in the forged output.

\begin{figure}[]
      \centering
      \includegraphics[width=\linewidth]{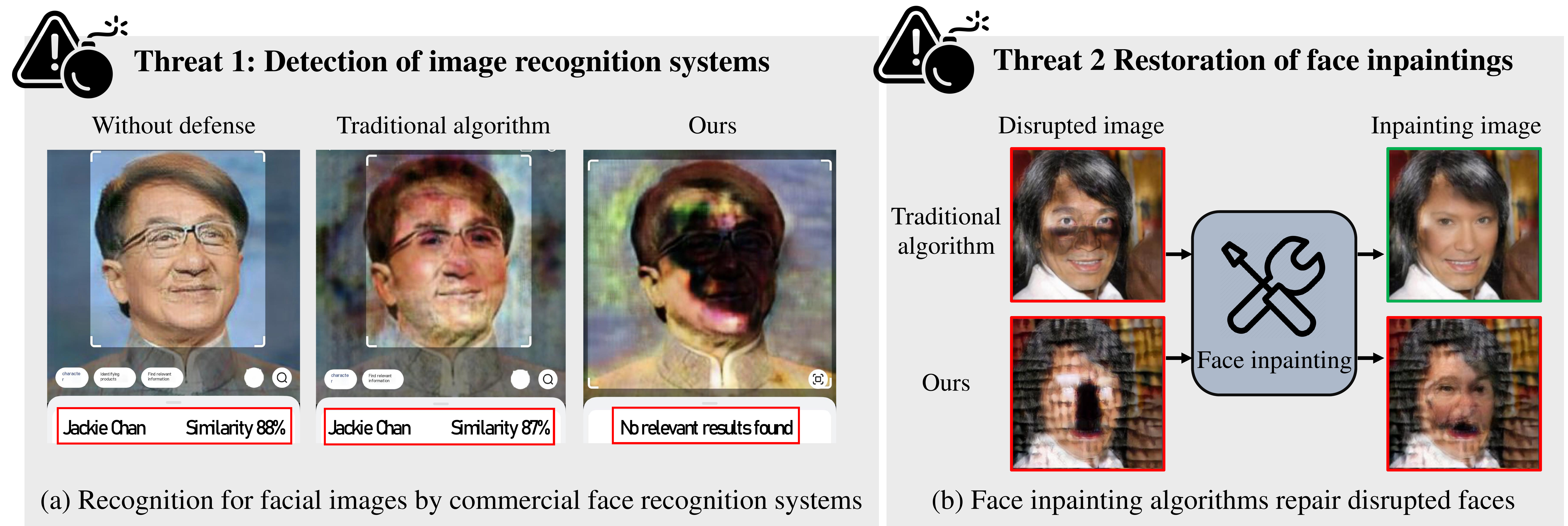}
      \caption{Illustration of potential threats to the insufficiently disrupted facial example. Challenges come primarily from commercial facial recognition systems and facial inpainting algorithms.}
      \label{fig:threat}
\end{figure}

Furthermore, as illustrated in Fig.~\ref{fig:threat}, inadequately disrupted facial images face two additional risks: 1) residual identifiable information increases the chance of recognition by commercial facial recognition systems, exacerbating stigmatization, especially since some entertainment applications automatically detect and promote celebrity images; 2) technically skilled malicious forgers may restore insufficiently distorted forged images using facial inpainting, enabling continued fraudulent activities.

\begin{figure}[]
      \centering
      \includegraphics[width=\linewidth]{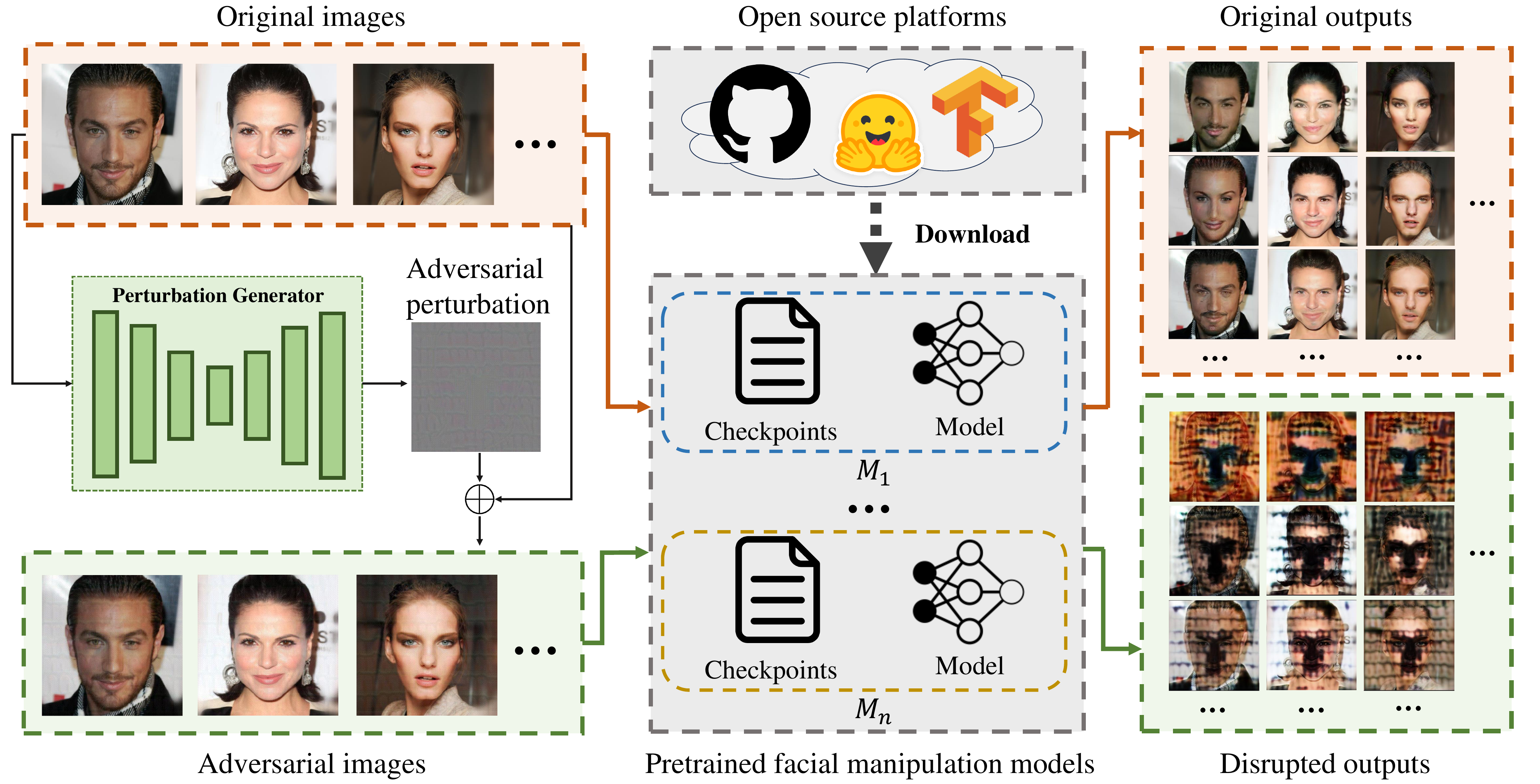}
      \caption{The publicly available pre-trained models can be easily downloaded from open-source platforms to implement forgeries. For a given image, the proposed ID-Guard can generate transferable perturbations for defense against multiple open-source facial manipulations through a single forward propagation of an image reconstruction network.}
      \label{fig:up}
\end{figure}

To address these concerns, we propose a proactive defense framework, ID-Guard. The framework generates transferable adversarial perturbations through a single forward pass of an image reconstruction network, effectively countering multiple open-source facial manipulation algorithms, as shown in Fig.~\ref{fig:up}. To eliminate identifiable semantic information in forged images and prevent identity spoofing, we incorporate a novel Identity Destruction Module (IDM). Unlike random or uncontrolled distortions, the IDM disrupts identity-related features in a structured manner, ensuring that the individual’s identity becomes unrecognizable.

The transferability of adversarial perturbations is essential in practical applications, as the facial manipulation methods used by forgers are often unknown and uncontrollable. To address this, a dynamic weighting strategy is introduced during training of the perturbation generator. Specifically,  the robustness of different facial manipulation models varies due to differences in architecture and design. Assigning equal weights to the adversarial loss of different models during perturbation generator training can bias the perturbations toward those that are easier to attack, degrading overall effectiveness. Therefore, we formulate adversarial attacks on multiple models as a multi-task learning problem, allowing loss weights to be dynamically adjusted during training to achieve balanced cross-model performance. In addition, a gradient prior perturbation strategy is introduced to improve training stability and accelerate convergence.

As demonstrated by the experiments later in this paper, the proposed ID-Guard effectively distorts identifiable regions in facial images manipulated by various open-source models, preventing both observers and facial recognition systems from identifying individuals and circumventing facial inpainting techniques. Furthermore, we show that ID-Guard can serve as a plug-and-play module in adversarial training for facial manipulation models, thereby enhancing their adversarial robustness. In summary, the main contributions of this work are as follows:
\begin{enumerate}
	\item
    	\label{step:class1} We propose a novel general adversarial perturbation generation framework, termed ID-Guard, to prevent facial manipulation from stigmatizing individuals. A single forward pass of the generator produces perturbations that effectively disrupt various facial manipulations. Moreover, this framework can be seamlessly integrated with other tasks.
	\item
    	\label{step:class2} To ensure complete disruption of manipulated images and prevent individual identification, we introduce an Identity Destruction Module (IDM). The IDM guides perturbations to specifically target identity-related semantic features, thereby alleviating issues of facial stigmatization and addressing concerns related to commercial facial recognition systems and image inpainting.
        \item
    	\label{step:class3} To enhance the transferability of generated adversarial perturbations, we formulate attacks on multiple facial manipulation models as a multi-task learning problem and design a dynamic weighting strategy. Additionally, a gradient prior perturbation strategy is proposed to improve the stability of the generator.
\end{enumerate}

The remainder of this paper is organized as follows: Related works on facial manipulation and proactive defense are reviewed in Section~\ref{Section II}. Section~\ref{Section III} presents the details of our method. The experimental results and analysis are provided in Section~\ref{Section IV}, followed by the conclusion in Section~\ref{Section V}.

\section{Related Works}  \label{Section II}
\subsection{Facial Manipulation}
Facial manipulation refers to the controlled modification of facial attributes in an image or video to produce desired visual content, such as identity, expression, age, and hair color. In recent years, leveraging the success of Generative Adversarial Networks (GANs) in image synthesis, numerous GAN-based algorithms~\cite{choi2018stargan,tang2021attentiongan,siddiquee2019learning,wu2019relgan,li2021image} with various architectures and constraints have been developed to enable facial manipulation~\cite{liu2023gan}. Some researchers have released their work as open-source on public platforms, providing pre-trained models and executable scripts, thereby greatly lowering the technical barriers for users to generate high-quality, high-fidelity fake images and videos.

\subsection{Proactive Defense against Facial Manipulation}
From the perspective of defense objectives, generalized proactive defense methods can be divided into two categories: proactive forensics and proactive disruption. Proactive forensics involves embedding imperceptible watermarks or traceable markers into multimedia content to aid in identifying manipulated samples. These embedded elements enable defenders to verify content authenticity and trace the origins of facial manipulations. In contrast, proactive disruption aims to degrade the quality of facial manipulation outputs by injecting adversarial perturbations, thereby misleading generative models. By distorting the generated results, proactive disruption effectively reduces the realism and credibility of forged images.

\subsubsection{Proactive Forensics}
Proactive forensic techniques aim to embed identifiable patterns into images to facilitate fake detection and manipulation provenance tracking. Early approaches, such as FaceGuard~\cite{yang2021faceguard} and Faketagger~\cite{wang2021faketagger}, detected forged examples by embedding watermarks into real images and verifying their integrity upon retrieval but lacked structured tracking mechanisms. To enable identity source tracking, Zhao \emph{et al.}\cite{zhao2023proactive} embedded watermarks as anti-Deepfake labels into facial identity features, allowing fake detection by verifying the presence of the label. These works primarily focus on authenticity detection but cannot localize tampered regions. To improve detection and localization, Asnani \emph{et al.} proposed a proactive embedding framework\cite{asnani2022proactive}, later refined into MALP~\cite{asnani2023malp}, which integrates attention mechanisms for fine-grained manipulation localization. PADL~\cite{bartolucci2024perturb} further enhanced robustness by combining perturbation-based defenses with detection and localization strategies. Zhao \emph{et al.}~\cite{zhao2024proactive} embedded a semi-fragile watermark in the original image, enabling localization of tampered regions by comparing retrieved and original watermarks upon counterfeiting. The limitation of proactive forensics is that it preserves the integrity of forged samples, whereas proactive disruption directly degrades or nullifies the forgery. Therefore, in this paper, we focus on proactive disruption methods.

\subsubsection{Proactive Disruption}
Recent studies have explored proactive disruption of facial manipulation by injecting adversarial perturbations into images. Ruiz~\emph{et al.}\cite{ruiz2020disrupting} and Yeh~\emph{et al.}\cite{yeh2020disrupting} disrupted facial manipulation by deriving gradient-based adversarial perturbations on target models. Subsequent works~\cite{qu2024df,guan2024adversarial} further enhanced the robustness of adversarial perturbations to protect personal images. However, due to structural and design disparities among facial manipulation models, perturbations crafted for a specific model often exhibit poor transferability to others, limiting their practical utility.
To mitigate this issue, studies such as~\cite{dong2023restricted,huang2021initiative} generated perturbations on surrogate models and transferred them to inaccessible models. Yet, substantial architectural discrepancies still hinder transfer effectiveness. A widely adopted alternative is model ensembling~\cite{huang2022cmua,aneja2022tafim,tang2023feature,zhu2023information}, which produces cross-model transferable adversarial perturbations and improves defense performance against diverse facial manipulation models to a certain extent.
However, these approaches overlook variations in adversarial robustness and gradient optimization across different models, resulting in inconsistent performance and an overall drop in defense effectiveness—a limitation addressed in this paper. Furthermore, they do not account for the issue of facial stigmatization caused by unconstrained perturbations. Zhai \emph{et al}. \cite{zhai2023defending} addressed this problem by embedding specific warning patterns into generated fake images, whereas our proposed ID-Guard directly distorts the facial recognition regions of fake images to disrupt identity inference.

\subsection{Multi-task Learning}
One of the effective routes to achieve multi-task learning is to dynamically weight the losses of different tasks according to their learning stages or the difficulty of learning. Sener \emph{et al}.~\cite{sener2018multi} pointed out that multi-task learning can be regarded as a multi-objective optimization problem, aiming to find the Pareto optimal solution to optimize the performance of multiple tasks. A representative method that has been proven to be effective and widely used is the multiple gradient descent algorithm (MGDA)~\cite{desideri2012multiple}. Some heuristics \cite{chen2018gradnorm,kendall2018multi,guo2018dynamic,liu2019end,xie2022end} measured the difficulty of a task based on the order of magnitude or change rate of the loss value, and then dynamically adjusted the weights of different tasks to obtain balanced performance. In this work, we further explore the potential of integrating multi-task learning strategies into cross-model transferable perturbation generation.

\section{Methodology}  \label{Section III}
In this section, the specific design and implementation details of the proposed ID-Guard framework are elaborated. For clarity, we first introduce an overview of the framework and a definition of notation.

%\clearpage
\subsection{Overview}
\subsubsection{Facial Manipulation}
Facial manipulation can be regarded as an image translation task, aiming to transform a given original example into a target manipulated example. Specifically, given an original face image $x \in \mathbb{R}^{3 \times H \times W}$, the facial manipulation model $\mathcal{M}$ leverages the specified target attribute or identity $a$ to map it to a forged image $y$. The process is formulated as:
\begin{equation}
 \begin{split}
    y = \mathcal{M}(x,a).
\end{split}
\end{equation}
Since end-to-end facial manipulation models employ different methods for embedding attributes or identity information, we simplify the manipulation process as $y=\mathcal{M}(x)$, where the face in the manipulated image $y$ retains the identity of the person in the original image $x$ while exhibiting the specified features of attribute $a$.

\subsubsection{Adversarial Perturbation against Manipulation}
To proactively disrupt facial manipulation, the objective of the defender is to generate an imperceptible adversarial perturbation $\delta$ such that the target facial manipulation model fails to map the adversarial image $x_{adv}=x+\delta$ to an acceptable manipulated output. The optimization of $\delta$ can be formulated as a maximization problem:
\begin{equation}
 \begin{split}
    &\max_{\delta} \mathcal{D}(\mathcal{M}(x),\mathcal{M}(x + \delta)),  \\
    &s.t. \quad \Vert \delta \Vert_{\infty} \le \epsilon,
\end{split}
\end{equation}

where $\epsilon$ is the infinite norm bound used to restrict the perturbation, and $\mathcal{D}$ is the distance metric between the original forged image $\mathcal{M}(x)$ and the disrupted forged image $\mathcal{M}(x + \delta)$. Some existing works~\cite{ruiz2020disrupting,yeh2020disrupting} employ gradient-based adversarial attack algorithms~\cite{goodfellow2014explaining,madry2017towards} to generate perturbations. However, these methods require multiple iterations for each perturbation generation, leading to high computational overhead. A more efficient approach is to train a perturbation generator $\mathcal{G}$, which can produce adversarial perturbations with a single forward pass during inference, i.e., $\delta=\mathcal{G}(x)$. The optimization problem for training $\mathcal{G}$ is defined as follows:
\begin{equation}
\label{eq2}
 \begin{split}
    &\max_{{\theta}_{\mathcal{G}}} \mathbb{E} (\mathcal{D}(\mathcal{M}(x),\mathcal{M}(x + \mathcal{G}(x)))),  \\
    &s.t. \quad \Vert \mathcal{G}(x) \Vert_{\infty} \le \epsilon,
\end{split}
\end{equation}
where ${\theta}_{\mathcal{G}}$ is the parameter of the perturbation generator. On one hand, previous state-of-the-art methods \cite{ruiz2020disrupting,yeh2020disrupting,huang2021initiative,huang2022cmua} typically apply the Mean Squared Error (MSE) loss as a proxy for the distance metric $\mathcal{D}$. However, this approach often results in unstructured distortions in the manipulated image while preserving identifiable facial features, leading to facial stigmatization. On the other hand, the perturbations generated by these methods generally lack universality and can only disrupt a single target model $\mathcal{M}$.

\begin{figure*}[]
      \centering
      \includegraphics[width=\linewidth]{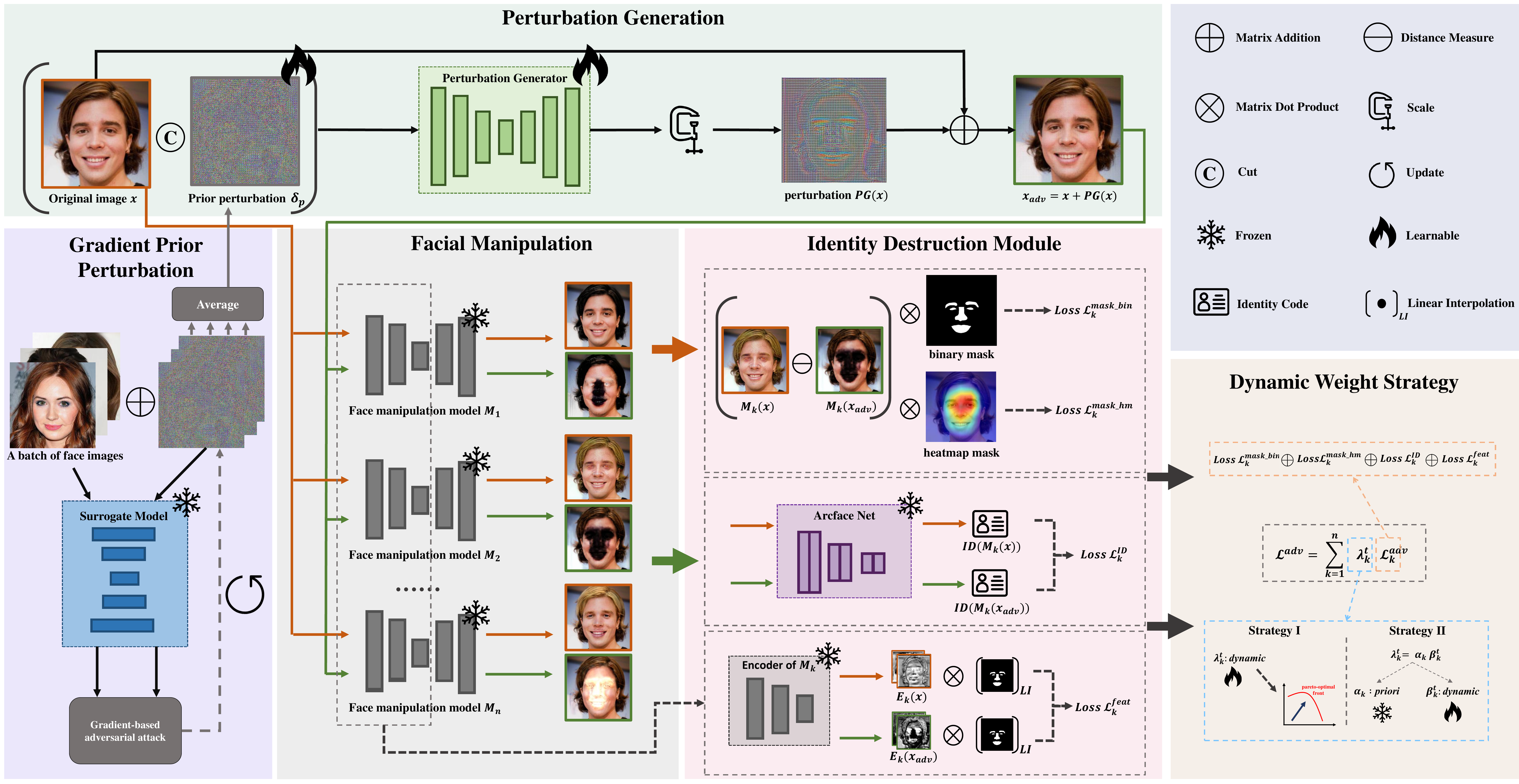}
      \caption{Illustration of the proposed ID-Guard framework. The perturbation generator takes a natural image $x$ as input and requires only one forward propagation to generate a cross-model adversarial perturbation dedicated to the input face that can be used to defend against multiple facial manipulations. In the training phase, ID-Guard consists of three modules, including the Identity Destruction Module, the dynamic weighting strategy, and the gradient prior perturbation strategy. The notation descriptions are shown in the upper right corner for reference.}
      \label{fig:framework}
\end{figure*}

\subsubsection{ID-Guard Framework}
To address the aforementioned issues, in this paper, we propose the ID-Guard framework, which aims to train a generator $\mathcal{G}$ with Resnet~\cite{he2016deep} architecture capable of producing the cross-model transferable adversarial perturbation, as shown in Fig.~\ref{fig:framework}. This generator produces customized perturbations for given images, effectively defending against a set of pre-trained facial manipulation models $\mathcal{S}_{\mathcal{M}} = \left\{ \mathcal{M}_{1},\mathcal{M}_{2},\dots,\mathcal{M}_{N} \right\}$. Following the model ensemble paradigm, for all accessible models $\mathcal{M}_{k}$, Eq.~\eqref{eq2} in our proposed framework can be rewritten as:
\begin{equation}
\label{eq3}
 \begin{split}
    &\max_{{\theta}_{\mathcal{G}}} \sum_{k=1}^N \lambda_{k} \mathbb{E} (\mathcal{D}(\mathcal{M}_{k}(x),\mathcal{M}_{k}(x + \mathcal{G}(x)))),  \\
    &s.t. \quad \Vert \mathcal{G}(x) \Vert_{\infty} \le \epsilon,
\end{split}
\end{equation}
where $\mathcal{S}_{\lambda} = \left\{ \lambda_{1},\lambda_{2},\dots,\lambda_{N} \right\}$ represents a set of weights used to balance the adversarial loss for different models during training. Determining these weights is one of our key tasks. To overcome the adversarial gradient discrepancies caused by significant structural differences among facial manipulation models, we propose a dynamic weighting strategy to achieve balanced cross-model defense performance, which will be introduced in Section~\ref{dynamic weighting strategy}. Additionally, to address the challenge of facial stigmatization caused by non-specific constraint distortions, we propose an Identity Destruction Module (IDM) to compute the distance metric $\mathcal{D}$, which will be detailed in Section~\ref{Identity Destruction Module}. Moreover, to incorporate structural adversarial gradient priors during the training of $\mathcal{G}$, we introduce a Gradient Prior Strategy, which is discussed in Section~\ref{Gradient Prior Perturbation}.

\subsection{Identity Destruction Module} \label{Identity Destruction Module}
To clarify the algorithmic design, we first introduce the Identity Destruction Module (IDM). Based on this module, we then provide a complete definition of the training loss function for the perturbation generator $\mathcal{G}$. As shown in Fig.~\ref{fig:framework}, the IDM consists of three sub-modules, which will be introduced separately next.

\subsubsection{Mask-Constrained Loss} First, we consider using face masks to limit the regions of image distortion by adversarial perturbations. The designed mask is two-fold: 1) the binary mask is used to restrict image distortion to areas of facial components including the eyes, nose, mouth, and eyebrows, which are proven to play an important role in identity recognition by human eyes~\cite{brunelli1993face,haxby2002human,singh2014face}; 2) the heatmap mask weights the face distortion loss at the pixel level, making the perturbation pay more attention to the important feature areas of the face. In this work, the heatmap of each image is obtained by solving Grad-cam~\cite{selvaraju2017grad} on VGGFace~\cite{simonyan2014very}. This design will also facilitate distorted images against commercial facial recognition systems. Hence, for the facial manipulation model $\mathcal{M}_{k}$, the mask-constrained loss can be formulated as:
\begin{equation}
\mathcal{L}_{k}^{mask\_bin} = \Vert \mathcal{M}_{k}(x) \odot m^{bin} - \mathcal{M}_{k}(x + \mathcal{G}(x)) \odot m^{bin} \Vert_{2},
\end{equation}
\begin{equation}
\mathcal{L}_{k}^{mask\_hm} = \Vert \mathcal{M}_{k}(x) \odot m^{hm} - \mathcal{M}_{k}(x + \mathcal{G}(x)) \odot m^{hm} \Vert_{2},
\end{equation}
where $m^{bin}$ and $m^{hm}$ denote the binary mask and heatmap mask of the original image $x$, respectively. Note that these masks are only computed during the perturbation generator training stage to constrain the distortion region and are not used in the inference process. $\odot$ indicates the element-wise multiplication.

\subsubsection{Identity Consistency Loss}
In addition to pixel-level constraints, we also consider maximizing the identity discrepancy between the forgery outputs of the facial manipulation model $\mathcal{M}_k$ for the original image and the adversarial image, i.e., maximizing the discrepancy between the original forged image $\mathcal{M}_k(x)$ and the disrupted forged image $\mathcal{M}_k(x_{adv})$. This is achieved by minimizing the cosine distance between their identity embeddings:
\begin{equation}
\mathcal{L}_{k}^{ID} = - \cos(ID( \mathcal{M}_{k}(x)),ID(\mathcal{M}_{k}(x + \mathcal{G}(x)))),
\end{equation}
where $ID(\cdot)$ denotes the Arcface net \cite{deng2022arcface} designed to extract high-quality features from facial images and embed them into a low-dimensional space where the distance between different embeddings corresponds to the similarity between faces.

\subsubsection{Feature Confusion Loss}
It is crucial to exploit the commonalities between different facial manipulation models during perturbation generation. Learning from~\cite{tang2023feature} that while the attribute embedding processes in end-to-end facial manipulation models vary, their feature extraction processes share similarities. In addition, feature-level perturbations can retain their effective components in network transmission to a greater extent~\cite{zhai2023defending}. Therefore, a feature confusion loss that enables the generator to focus on destroying feature-level faces is incorporated into the IDM to improve the effectiveness and transferability of the produced perturbation:
\begin{equation}
\mathcal{L}_{k}^{feat} = \Vert E_{k}(x) \odot LI(m^{bin}) - E_{k}(x + \mathcal{G}(x)) \odot LI(m^{bin}) \Vert_{2},
\end{equation}
where $E_{k}$ is the feature extraction module of $\mathcal{M}_{k}$, which is defined here as the upsampling network of each model. $LI(\cdot)$ represents the linear interpolation operation to make the binary mask and the extracted feature map consistent in sample size.

In summary, given a pre-trained facial manipulation model $\mathcal{M}_{k} \in \mathcal{S}_{\mathcal{M}}$, the adversarial loss against it can be formulated as:
\begin{equation}
\mathcal{L}_{k}^{adv} = \mathcal{L}_{k}^{mask\_bin} + \mathcal{L}_{k}^{mask\_hm} + \omega_{1}\mathcal{L}_{k}^{ID} + \omega_{2}\mathcal{L}_{k}^{feat},
\end{equation}
where $\omega_{1}$ and $\omega_{2}$ are weighting factors that balance the identity consistency loss and the feature confusion loss. The loss function for training the perturbation generator $\mathcal{G}$ is a linear combination of adversarial losses of facial manipulation models:
\begin{equation}
\label{eq9}
\begin{split}
\mathcal{L}^{adv} &= \lambda_{1} \cdot \mathcal{L}_{1}^{adv} + \lambda_{2} \cdot \mathcal{L}_{2}^{adv} + \dots + \lambda_{N} \cdot \mathcal{L}_{N}^{adv} \\
&= \sum_{k=1}^{N} \lambda_k \mathcal{L}_k^{adv}.
\end{split}
\end{equation}
The definition of $\lambda_{k}$ is shown in Eq.~\eqref{eq3}, and the use of the proposed dynamic weighting strategy to solve it will be introduced next.

\subsection{Dynamic Weighting Strategy} \label{dynamic weighting strategy}
 We regard the attacks against different facial manipulation models in Eq.~\eqref{eq9} as a multi-task learning problem to optimize the weight set $\mathcal{S}_{\lambda}$, ensuring that the generated perturbations achieve more balanced cross-model adversarial effectiveness. As shown in Fig. \ref{fig:framework}, the proposed dynamic weighting strategy is two-fold.

\subsubsection{MGDA-based}
First, we integrate the Multiple Gradient Descent Algorithm (MGDA)~\cite{desideri2012multiple} into the proposed ID-Guard. The goal of MGDA is to find a set of weights $\lambda_{k}$ that balances the trade-offs between tasks while respecting the constraints of Pareto optimality. It can be formulated as the following quadratic programming problem:
\begin{equation}
    \begin{split}
        & \min_{\lambda \geq 0} \quad \left\| \sum_{k=1}^{N} \lambda_k \mathcal{L}_k^{adv} \right\|^2, \\
        & s.t. \quad \sum_{k=1}^{N} \lambda_i = 1.
    \end{split}
\end{equation}

In this paper, we employ the Frank-Wolfe method~\cite{sener2018multi} to solve the optimization problem. The core idea is to combine gradients from multiple tasks into a suitable descent direction that minimizes the losses across all tasks. Specifically, we follow the implementation outlined in \cite{sener2018multi}~\footnote{https://github.com/isl-org/MultiObjectiveOptimization}. This dynamic weighting strategy, based on MGDA, is used as a baseline, referred to as Strategy I (S-I).

\subsubsection{KPI-based}
Additionally, a dynamic weighting strategy based on Key Performance Indicators (KPI) is introduced, denoted as Strategy II (S-II). Each dynamic weight $\lambda_{k}$ is refined into a prior weight and a learnable weight, as follows:
\begin{equation}
\lambda_{k}^{t} = \alpha_{k} \times \beta_{k}^{t},
\end{equation}
where $t$ indexes the iteration steps, $\alpha_{k}=10^n \, (n \in \mathbb{Z})$ is the prior weight representing the magnitude of the attack loss, which reflects the adversarial robustness of different forged models, and $\beta_{k}^{t}$ is the learnable weight updated adaptively based on defense performance during each iteration.

To set the prior weight $\alpha_{k}$ reasonably, a small-sample heuristic method is used to quantify the prior adversarial robustness of different facial manipulation models. Specifically, a small batch of face images is randomly selected, and slight attacks are applied to each model using PGD~\cite{madry2017towards} with the same settings. The magnitude of the $L_2$ distance is used to determine the value of the prior weight. This approach provides a reasonable initialization of weights in the early stages of training, balancing the contribution of adversarial losses across different models.

For the learnable weight $\beta_{k}^{t}$, inspired by the study~\cite{guo2018dynamic}, we assign higher loss weights to more difficult-to-learn tasks, i.e.,  facial manipulation models that are harder to disrupt during training. The $\beta_{k}^{t}$ is computed as follows:
\begin{equation}
    \begin{split}
        & \beta_{k}^{t+1} = -(1- \mathcal{K}_{k}^{t}) \log \mathcal{K}_{k}^{t}, \\
        & \beta_{k}^{0} = 1,
    \end{split}
\end{equation}
where $\mathcal{K}_{k}^{t}$ is the KPI representing the attack difficulty of model $\mathcal{M}_{k}$ at iteration $t$. A higher KPI indicates that the model is easier to attack, so a smaller weight $\beta_{k}^{t}$ is assigned, while a lower KPI $\mathcal{K}_{k}$ results in a larger weight to enhance the adversarial impact of the generated perturbation against the model $\mathcal{M}_{k}$. Thus, the KPI value is inversely proportional to the adversarial loss weight. A negative correlation function of $\mathcal{K}_{k}^{t}$ is used to calculate the $\beta_{k}^{t}$.
For convenience, we use the proposed $\mathcal{L}_{k}^{mask\_hm}$ to compute the distance between the real and perturbed fake results as a proxy for KPI, with its range truncated to $(0,1]$ to ensure the monotonicity of the function. 
Unlike the MGDA-based dynamic weighting strategy, which requires additional computation in each iteration, the KPI-based strategy directly uses the loss function value calculated in the previous iteration as the KPI to obtain the learnable weight for this iteration. This combination design is more efficient and enhances the stability of the dynamic weight.

\subsection{Gradient Prior Perturbation}  \label{Gradient Prior Perturbation}
One obstacle to training adversarial perturbation generators is their lack of initial awareness of structural perturbation information. Therefore, we introduce a gradient prior perturbation strategy. Motivated by \cite{aneja2022tafim}, we consider jointly optimizing for a global prior perturbation $\delta_{p} \in \mathbb{R}^{3 \times H \times W}$ and the generator $\mathcal{G}$. Specifically, we first train a surrogate model $\mathcal{M}_{s}$ with face reconstruction capabilities, treating it as an approximate task of the facial manipulation~\cite{dong2023restricted,he2022defeating}. Next, we use PGD~\cite{madry2017towards} to derive gradient-based adversarial perturbations against $\mathcal{M}_{s}$ on a batch of face images, and average these perturbations to obtain $\delta_{p}$. More details will be introduced in \ref{Implementation Details}. Therefore, the overall optimization objective in Eq.~\eqref{eq3} can be rewritten as:
\begin{equation}
 \begin{split}
    \max_{{\theta}_{\mathcal{G}},\delta_{p}} \sum_{k=1}^N \lambda_{k} \mathbb{E}
    &(\mathcal{D}(\mathcal{M}_{k}(x),\mathcal{M}_{k}(x + \mathcal{G}(\mathrm{cat}(x,\delta_{p})))) +
    \\
    &\mathcal{D}(\mathcal{M}_{k}(x),\mathcal{M}_{k}(x +\delta_{p})  ),  \\
    &s.t. \quad \Vert \mathcal{G}(x) \Vert_{\infty} \le \epsilon, \quad \Vert \delta_{p} \Vert_{\infty} \le \epsilon,
\end{split}
\end{equation}
where $\mathrm{cat}(\cdot)$ denotes channel-wise concatenation, i.e., $\mathrm{cat}(x,\delta_{p}) \in \mathbb{R}^{6 \times H \times W}$. Both parts are calculated using the adversarial loss defined in Eq.~\ref{eq9}. The intuition behind this design is that the gradient prior perturbation provides the generator with rich information about the prior gradient and perturbation structure, thereby promoting more stable training and more efficient perturbation generation. In particular, the optimized prior adversarial perturbation $\delta_{p}$ (noted as P-Pert) can be viewed as a cross-model universal adversarial perturbation that can protect multiple images from multiple facial manipulation models. The complete training process is given in Algorithm~\ref{algorithm1}.

\begin{algorithm}[ht]
    \SetKwInOut{Input}{input}\SetKwInOut{Output}{output}
    \Input{Original dataset $S_{ori}$, facial
manipulation model set $\mathcal{S}_{\mathcal{M}}$, max iteration $T$.}
    \Output{Optimized $\mathcal{G}*$ and $\delta_{p}*$.}
    
    \text{// Generating initial prior adversarial perturbations $\delta_{p}^0$}\;
    $\delta_{p}^0 \leftarrow 0$\; 
    $\hat{S}_{ori} \leftarrow SelectBatch(S_{ori})$\;  
    \ForEach{$x^{i} \in \hat{S}_{ori}$}
    {
        \text{// Implementing the PGD}\;
        $\delta^{i} \leftarrow PGD(x^{i})$\;
        $\delta_{p}^0 \leftarrow \delta_{p}^0 + \delta^{i}$ \;
        
    }

    $\delta_{p}^0 \leftarrow \delta_{p}^0 / Len(\hat{S}_{ori})$ \;

    $\mathcal{G}_{0} \leftarrow RandomInit()$\;
    \ForEach{$\lambda_{k} \in \mathcal{S}_{\lambda}^{0}$}
    {
        $\lambda_{k}^{0} \leftarrow 1$\;
    }
    \text{// Training the Perturbation Generator}\;
    \For{$j\leftarrow1$ \KwTo $T$}
    {   
        $x_b \leftarrow SelectBatch(S_{ori})$\;
        \text{// Calculating the prior weight set}\;
        $\mathcal{S}_{\lambda}^{j} \leftarrow Strategy(\mathcal{G}^{j-1},\delta_{p}^{j-1},x_b,\mathcal{L}^{adv},\mathcal{S}_{\mathcal{M}})$\;
        \text{// Updating the Parameters}\;
        $\mathcal{G}^{j},\delta_{p}^{j}  \leftarrow Update(\mathcal{G}^{j-1},\delta_{p}^{j-1},x_b,\mathcal{L}^{adv},\mathcal{S}_{\lambda}^{j},\mathcal{S}_{\mathcal{M}})$\; 
    }
    $\mathcal{G}* \leftarrow \mathcal{G}^{N}$\;
    $\delta_{p}* \leftarrow \delta_{p}^{N}$\;
    \textbf{Return} {$\mathcal{G}*,\delta_{p}*$}
    \caption{Training of the Perturbation Generator}
    \label{algorithm1}
\end{algorithm}

\subsection{ID-Guard for Adversarial Training} \label{AT}
A potential application of ID-Guard is leveraging the proposed perturbation generator $\mathcal{G}$ as an adversarial attack module for adversarial training of facial manipulation models to enhance their robustness. During adversarial training, the well-trained $\mathcal{G}$ can rapidly generate training images with adversarial patterns, which are then used to train the facial manipulation model $\mathcal{M}$. Meanwhile, $\mathcal{G}$ is simultaneously fine-tuned to adapt to $\mathcal{M}$.

Specifically, adversarial training can be formulated as a bi-level min-max optimization problem, where the inner maximization aims to generate adversarial examples that maximize the loss, while the outer minimization seeks to update the model parameters to minimize the bad-case loss. Given a facial manipulation model $\mathcal{M}$, it can be formulated as:
\begin{equation}
\min_{{\theta}_\mathcal{M}} \left[ \max_{{\theta}_\mathcal{G},\mathcal{G}(x) \in \mathcal{S}} \mathcal{L}(\mathcal{M}(x + \mathcal{G}(x)), y) \right],
\end{equation}
where, the adversarial perturbation $\delta = \mathcal{G}(x)$ is constrained within a \(\ell_\infty\)-bounded perturbation set $\mathcal{S}$, \textcolor{blue}{and y is the desired output}. The inner maximization problem optimizes the perturbation generator $\mathcal{G}$ to produce adversarial perturbations that maximize the loss. Specifically, the adversarial loss for the selected facial manipulation model $\mathcal{M}$ is computed in this step. The outer minimization then updates the parameters of $\mathcal{M}$ to enhance its robustness against such perturbations. Here, the training loss of $\mathcal{M}$ defined in the specific algorithm is computed. We will discuss its effectiveness in Section~\ref{ID-Guard for Adversarial Training}.

\section{Experiments}  \label{Section IV}
\subsection{Experimental Setup}
\subsubsection{Datasets}
In our experiments, the CelebAMask-HQ dataset~\cite{lee2020maskgan} is selected to train the perturbation generator. It consists of more than 30,000 face images, where each image carries semantic masks for 19 facial component categories. These fine-grained mask labels can provide support for computing the binary mask-constrained loss during the training stage. To adequately evaluate the performance as well as the generalizability of our method and the competing algorithms, we test them on three datasets, including CelebAMask-HQ~\cite{lee2020maskgan}, LFW~\cite{huang2008labeled}, and FFHQ~\cite{karras2019style}.

\begin{figure}[t]
      \centering
      \includegraphics[width=\linewidth]{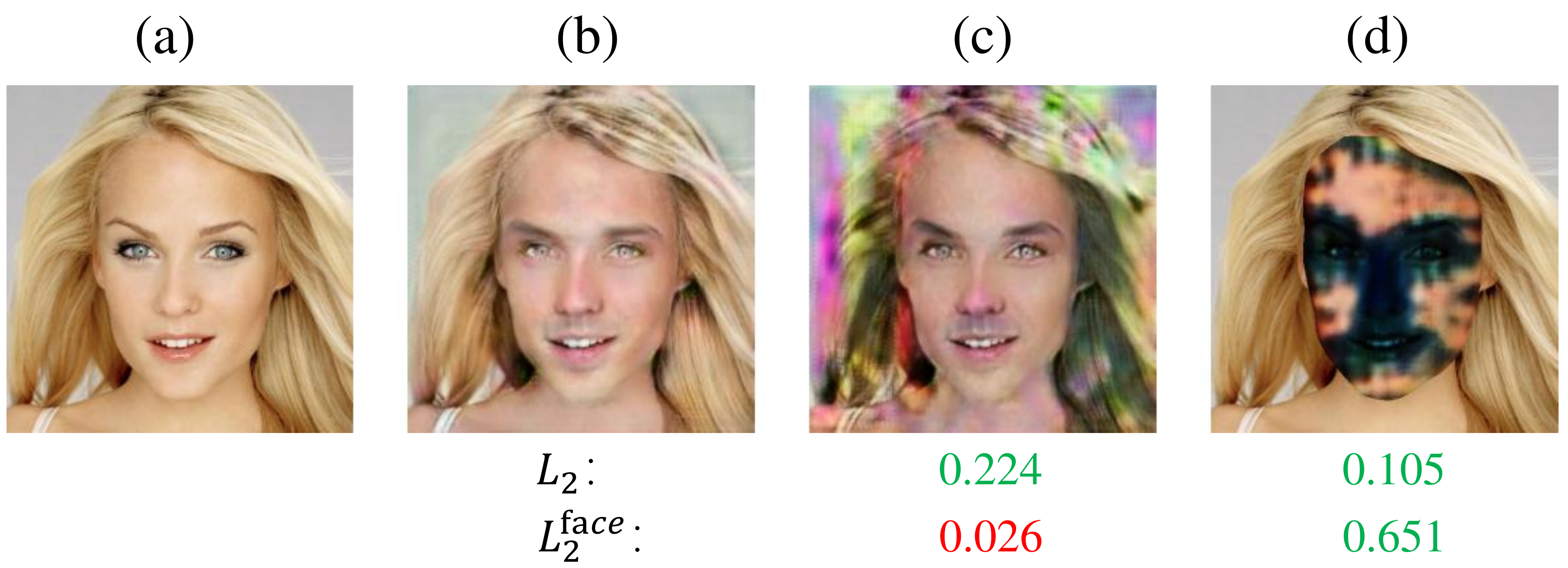}
      \caption{Visual example of $L_{2}^{face}$ metric design. (a) is a natural image, (b) is a forged image, and (c) and (d) are disrupted images in two different situations.}
      \label{fig:l2face}
\end{figure}

\begin{figure}[]
      \centering
      \includegraphics[width=\linewidth]{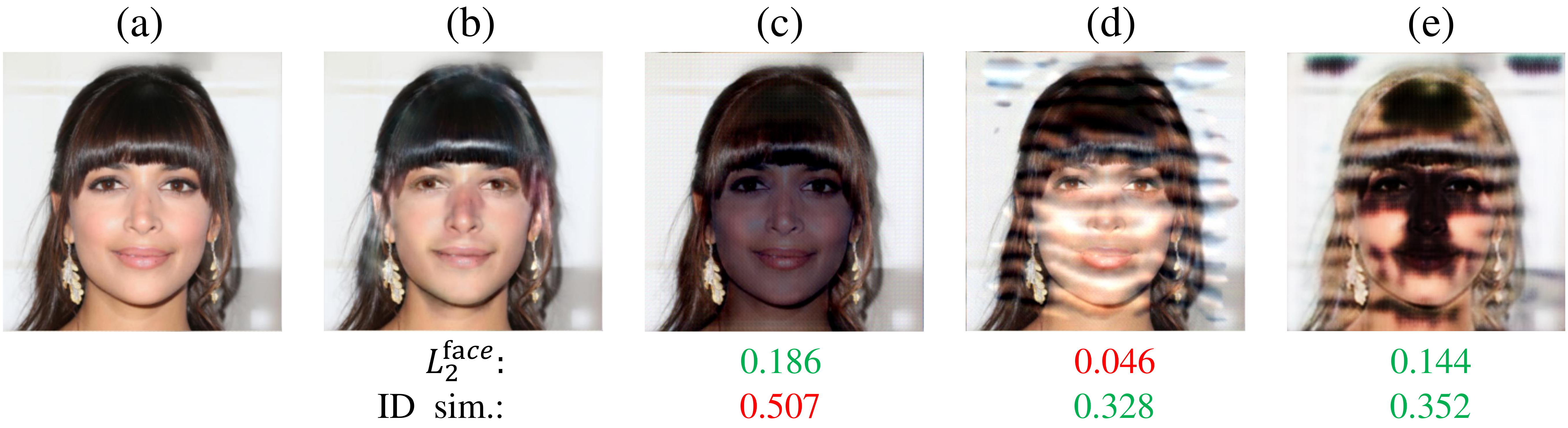}
      \caption{Illustration of indicators that determine the success of a defense. (a) is a natural image, (b) is a forged image, (c), (d), and (e) are distorted images under different defense situations, respectively. It can be seen that when only one of the metrics, $L_{2}^{face}$ distance or identity similarity, satisfies the set conditions, it is not sufficient to break the identity of the individual in the image.}
      \label{fig:eval}
\end{figure}

\subsubsection{Target Models}
We choose five facial manipulation models, including StarGAN \cite{choi2018stargan}, AGGAN~\cite{tang2021attentiongan}, FPGAN~\cite{siddiquee2019learning}, RelGAN~\cite{wu2019relgan}, and HiSD~\cite{li2021image} as target models to implement the attack, and they are all trained on the CelebA~\cite{liu2015deep} dataset. In the experiment, for StarGAN, AGGAN, FPGAN, and RelGAN, we select black hair, blond hair, brown hair, gender, and age as editing attributes; for HiSD, five images with black hair, blond hair, brown hair, glasses, and bangs are chosen as attribute references, respectively.

\begin{table*}[]
    \renewcommand{\arraystretch}{1.3}
    \centering
    \caption{Quantitative comparison for disrupting different target models. For each column within the same dataset. The best result is marked in \textbf{bold}, while the second best result is marked with an \underline{underline}. }
    \label{tab:mainresult}
    
    \resizebox{2.0\columnwidth}{!}{
        \begin{tabular}{c|c|ccccccccccccccc}
        \toprule
        \multirow{2}{*}{Datasets} 
        &\multirow{2}{*}{Methods} 
        &\multicolumn{3}{c}{StarGAN\cite{choi2018stargan}}  &\multicolumn{3}{c}{AGGAN\cite{tang2021attentiongan}} 
        &\multicolumn{3}{c}{FPGAN\cite{siddiquee2019learning}} 
        &\multicolumn{3}{c}{RelGAN\cite{wu2019relgan}}
        &\multicolumn{3}{c} {HiSD\cite{li2021image}} \\
      \cmidrule(l){3-5}  \cmidrule(l){6-8}  \cmidrule(l){9-11} \cmidrule(l){12-14} \cmidrule(l){15-17} & &$L_{2}^{face}$$\uparrow$ &ID sim.$\downarrow$ &DSR$\uparrow$ &$L_{2}^{face}$$\uparrow$ &ID sim.$\downarrow$ &DSR$\uparrow$ &$L_{2}^{face}$$\uparrow$ &ID sim.$\downarrow$ &DSR$\uparrow$ &$L_{2}^{face}$$\uparrow$ &ID sim.$\downarrow$ &DSR$\uparrow$ &$L_{2}^{face}$$\uparrow$ &ID sim.$\downarrow$ &DSR$\uparrow$\\
        \midrule
        
        \multirow{7}{*}{CelebAMask-HQ} 
        &Disrupting\cite{ruiz2020disrupting}           
        &\textbf{1.047} &\underline{0.023} &\textbf{1.000} &0.114 &0.479 &0.292 &0.134 &0.369 &0.472 &0.021 &0.753 &0.001 &0.004 &0.839 &0.001\\
        &PG\cite{huang2021initiative}          
        &0.101 &0.302 &0.646 &0.032 &0.658 &0.014 &0.069 &0.352 &0.458 &0.007 &0.836 &0.005 &0.016 &0.617 &0.095\\
        &CMUA\cite{huang2022cmua}            
        &0.586 &0.368 &0.584 &0.062 &0.646 &0.016 &0.052 &0.486 &0.126 &\underline{0.296} &0.630 &0.070 &0.055 &0.603 &0.165\\
        &IAP\cite{zhu2023information}             
        &0.450 &0.118 &\underline{0.994} &0.054 &0.300 &\underline{0.398} &0.321 &0.181 &\underline{0.928} &0.109 &0.545 &0.165 &0.056 &0.193 &0.590\\
        \cmidrule(lr){2-17} 
        &Ours (S-I)       
        &0.362 &0.055 &\textbf{1.000} &\textbf{0.376} &\textbf{0.062} &\textbf{1.000} &\underline{0.558} &\underline{0.004} &\textbf{1.000} &0.285 &\underline{0.065} &\underline{0.998} &\textbf{0.205} &\textbf{0.018} &\textbf{1.000}\\
        &Ours (S-II)      
        &\underline{0.592} &\textbf{0.016} &\textbf{1.000} &\underline{0.298} &\underline{0.088} &\textbf{1.000} &\textbf{0.635} &\textbf{0.001} &\textbf{1.000} &\textbf{0.404} &\textbf{0.013} &\textbf{1.000} &\underline{0.204} &\underline{0.043} &\underline{0.998}\\
        &Ours (P-Pert)
        &0.246 &0.031 &\textbf{1.000} &0.288 &0.078 &\textbf{1.000}
        &0.551 &0.009 &\textbf{1.000} &0.069 &0.176 &0.770 &0.136 
        &0.171 &0.902 \\
        \midrule
        
        \multirow{7}{*}{LFW} 
        &Disrupting\cite{ruiz2020disrupting}           
        &\textbf{0.956} &0.062 &\textbf{1.000} &0.142 &0.443 &0.412 &0.126 &0.380 &0.546 &0.023 &0.075 &0.011 &0.004 &0.849 &0.000\\
        &PG\cite{huang2021initiative}          
        &0.134 &0.306 &0.728 &0.053 &0.656 &0.044 &0.069 &0.415 &0.356 &0.008 &0.838 &0.002 &0.020 &0.651 &0.035\\
        &CMUA\cite{huang2022cmua}            
        &0.513 &0.247 &0.788 &0.092 &0.462 &0.294 &0.054 &0.745 &0.108 &0.231 &0.618 &0.078 &0.063 &0.600 &0.150\\
        &IAP\cite{zhu2023information}             
        &0.413 &0.411 &\underline{0.956} &0.085 &0.411 &\underline{0.424} &0.310 &0.161 &0.946 &0.079 &0.533 &0.211 &0.071 &0.238 &0.545\\
        \cmidrule(lr){2-17}
        &Ours (S-I)        
        &0.328 &\textbf{0.021} &\textbf{1.000} 
        &\textbf{0.446} &\textbf{0.033} &\textbf{1.000} &\underline{0.555} &\underline{0.027} &\underline{0.994} &\underline{0.271} &\underline{0.084} &\underline{0.992} &\underline{0.178} &\textbf{0.053} &\textbf{0.998}\\
        &Ours (S-II)     
        &\underline{0.529} &\underline{0.051} &\textbf{1.000} &\underline{0.417} &\underline{0.068} &\textbf{1.000} &\textbf{0.646} &\textbf{0.017} &\textbf{1.000} &\textbf{0.429} &\textbf{0.025} &\textbf{1.000} &\textbf{0.226} &\underline{0.078} &\underline{0.987}\\
        &Ours (P-Pert)
        &0.192 &0.052 &\textbf{1.000} 
        &0.411 &0.070 &\textbf{1.000}
        &0.535 &0.029 &\textbf{0.991} 
        &0.074 &0.228  &0.767 
        &0.175 &0.238 &0.845 \\
        \midrule
        
        \multirow{7}{*}{FFHQ} 
        &Disrupting\cite{ruiz2020disrupting}           
        &\textbf{0.956} &\underline{0.033} &\textbf{1.000} &0.142 &0.487 &0.312 &0.126 &0.409 &0.426 &0.023 &0.747 &0.013 &0.005 &0.878 &0.002\\
        &PG\cite{huang2021initiative}          
        &0.134 &0.328 &0.628 &0.053 &0.708 &0.002 &0.069 &0.424 &0.360 &0.008 &0.839 &0.000 &0.020 &0.683 &0.108\\
        &CMUA\cite{huang2022cmua}            
        &0.515 &0.346 &0.624 &0.092 &0.635 &0.028 &0.054 &0.512 &0.168 &0.231 &0.666 &0.043 &0.063 &0.656 &0.120\\
        &IAP\cite{zhu2023information}             
        &0.413 &0.167 &\underline{0.976} &0.085 &0.365 &0.476 &0.310 &0.207 &\underline{0.898} &0.079 &0.568 &0.123 &0.071 &0.252 &0.525\\
        \cmidrule(lr){2-17}
        &Ours (S-I)        
        &0.328 &0.053 &\textbf{1.000} 
        &\textbf{0.446} &\textbf{0.089} &\textbf{0.994} &\textbf{0.557} &\textbf{0.019} &\textbf{1.000} &\underline{0.271} &\underline{0.112} &\underline{0.973} &\underline{0.178} &\textbf{0.050} &\textbf{0.965}\\
        &Ours (S-II)       
        &\underline{0.534} &\textbf{0.005} &\textbf{1.000} &\underline{0.330} &0.103 &0.970 &\underline{0.509} &\underline{0.023} &\textbf{1.000} &\textbf{0.348} &\textbf{0.028} &\textbf{0.995} &\textbf{0.184} &\underline{0.082} &\underline{0.963}\\
        &Ours (P-Pert)
        &0.213 &0.052 &\textbf{1.000} 
        &0.301 &\underline{0.090} &\underline{0.978} 
        &0.455 &0.046 &\textbf{1.000} 
        &0.063&0.246  &0.637 
        &0.118 &0.262 &0.730 \\
        \bottomrule
    \end{tabular}}
\end{table*}

\subsubsection{Baselines}
To demonstrate the superiority of the proposed method in face identity protection and cross-model transferable performance, four advanced proactive defense methods, including Disrupting \cite{ruiz2020disrupting}, PG~\cite{huang2021initiative}, CMUA~\cite{huang2022cmua}, and IAP~\cite{zhu2023information}, are selected as competing algorithms. Disrupting~\cite{ruiz2020disrupting} disrupts facial manipulation by iteratively solving gradient-based adversarial perturbations on the target model. 
PG~\cite{huang2021initiative} achieves transferable adversarial perturbation generation in gray-box scenarios by attacking a surrogate model. CMUA~\cite{huang2022cmua} is a baseline of universal defense against multiple models. IAP~\cite{zhu2023information} designs an information-containing adversarial perturbation. For a fair comparison, we implement only its proactive disruption component, while the information embedding and extraction aspects are discussed in Section~\ref{Further Discussion}.

\subsubsection{Metrics}
Unlike traditional evaluation methods that calculate the $L_{2}$ distance of the whole image or the forged area between the forged and distorted outputs, we focus on measuring the difference in the facial area of the output. Specifically, we introduce $L_{2}^{face}$, which can better reflect whether the defense successfully destroys the identity information of the face image, making it unrecognizable. $L_{2}^{face}$ can be expressed as:
\begin{equation}
L_{2}^{face}(y,\hat{y}) =  \frac{\sum_{i}\sum_{j}Face_{i,j}\odot(y_{i,j} - \hat{y}_{i,j})^{2}}{\sum_{i}\sum_{j}Face_{i,j}}
\end{equation}
where $(i,j)$ is the coordinate of pixels and $Face_{i,j}$ is the binary facial mask of the image. The pixel value of its face area is $1$; otherwise, it is $0$. The binary facial mask is calculated by Dlib~\footnote{https://pypi.org/project/dlib}.
The intuition behind this design is that the traditional full-image $L_2$ distance metric fails to reflect distortions in the facial region. As illustrated in Fig. \ref{fig:l2face}, although the disrupted image in (c) is reported as successful under the $L_2$ distance metric, the perturbation primarily affects the background while leaving the facial region largely intact. In contrast, the proposed $L_{2}^{face}$ metric provides a more accurate measure of identity-related distortions in the facial region, which aligns with human perception.
Additionally, we evaluate the identity similarity (noted as ID sim. in Tables) computed by Arcface~\cite{deng2022arcface} between the forged and distorted outputs. Defense success rates are also considered. Previous works \cite{ruiz2020disrupting,huang2022cmua} have generally determined the success of a defense by whether the $L_{2}$ distance is greater than $0.05$, but this is incomplete in the task of preventing face stigmatization. As shown in Fig.~\ref{fig:eval}, the distorted output in Fig.~\ref{fig:eval} (c) reports a successful defense at the $L_{2}^{face}$ distance, but it seems to only \textit{blacken} the face without destroying the individual's identity. Therefore, we propose that both $L_{2}^{face}$ distance greater than $0.05$ and identity similarity less than $0.4$ be satisfied to indicate a successful defense, which is a more challenging evaluation. The contrast between (d) and (e) in Fig.~\ref{fig:eval} shows the necessity of considering both restrictions simultaneously.

\subsubsection{Implementation Details} \label{Implementation Details}
All images used in the experiments are resized to a resolution of $256\times256$ and the pixel value is normalized to $\left[-1,1\right]$. For fairness, the bound $\epsilon$ of all competitive algorithms is restricted to $0.05$ to ensure the invisibility of the perturbation. For StarGAN, AGGAN, FPGAN, RelGAN, and HiSD, we set the prior weight $\alpha$ to $\left[1,10,1,10,100\right]$, respectively, as determined by a simple pre-experiment on the gradient-based adversarial attack against them. We derive the gradient prior perturbation on 2,000 randomly selected face images from CelebAMask-HQ~\cite{lee2020maskgan}, running the PGD~\cite{madry2017towards} for 10 iterations with a $0.01$ step size. The perturbation generator is trained using the Adam~\cite{kingma2014adam} with a learning rate of $0.0001$. The experiment was performed on a server running an Ubuntu 18.04.6 LTS system with an Intel(R) Core(TM) i9-10940X CPU @ 3.30 GHz and four NVIDIA 3090Ti GPUs with 24 GB of memory.

\begin{figure}[]
      \centering
      \includegraphics[width=\linewidth]{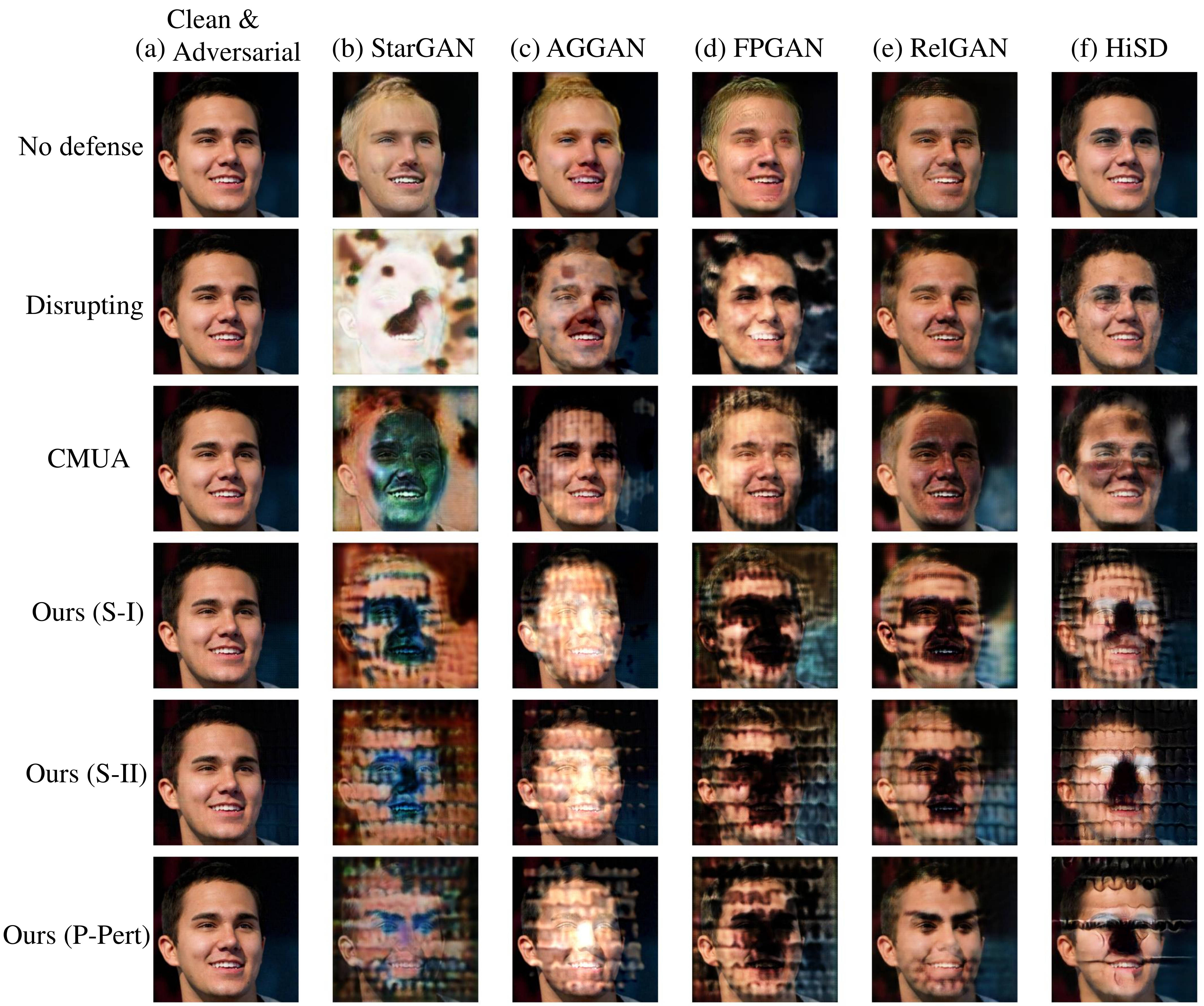}
      \caption{Visual examples of disruption to different facial manipulations.}
      \label{fig:result}
\end{figure}
 
\subsection{Comparison with Baselines}
Table~\ref{tab:mainresult} summarizes the quantitative comparison of the proposed ID-Guard with competitive algorithms. Our method is reported separately under two strategies, as presented in \ref{dynamic weighting strategy}. The Disrupting \cite{ruiz2020disrupting} produces gradient-based adversarial perturbations against StarGAN \cite{choi2018stargan}, which makes it overfit in disrupting this model at the cost of cross-model performance, as shown in Fig.~\ref{fig:result}. The perturbation optimization of CMUA~\cite{huang2022cmua} and PG~\cite{huang2021initiative} is unconstrained and thus has limited destruction to the identity semantics. The visual example of the CMUA in Fig.~\ref{fig:result} shows that the person's identity in the distorted image can still be recognized. Attributed to the feature correlation measurement loss employed, IAP~\cite{zhu2023information} improves the performance to destroy identification to some extent. In comparison, our method significantly destroys the identity semantics of images and effectively prevents the stigmatization of faces. Furthermore, the baseline methods equally weight the attack loss of different facial manipulations, leading to perturbations biased towards vulnerable models. Due to the introduction of the dynamic parameter strategy, we achieve balanced performance against various facial manipulations. For the most robust model against attack, HiSD~\cite{li2021image}, ID-Guard improves the defense success rate by $50\%$ compared with the state-of-the-art method.

On the three selected datasets, ID-Guard under both strategies demonstrated superior performance. For strategy I, the MGDA algorithm is employed to adjust the weights, which has the advantage of eliminating the need for human intervention and prior knowledge during the training process. However, this non-intervention makes MGDA-based ID-Guard exhibit an \textit{extreme effect} to some extent: it performs better on the more vulnerable model (e.g., AGGAN~\cite{tang2021attentiongan}) and the more robust model (e.g., HiSD~\cite{li2021image}), but does not provide significant improvement on models in the middle (e.g., RelGAN~\cite{wu2019relgan}). Moreover, additional backpropagation computation is required at each iteration when implementing MGDA, which increases the training overhead. In contrast, the KPI-based strategy balances different models by maintaining a set of prior parameters, ensuring more stable defense performance across models and a more balanced performance improvement. However, its drawback is that preliminary experiments are required before formal training to determine the values of the prior weight set.

We conducted an additional evaluation of the optimized gradient-prior perturbation, P-Pert. Similar to CMUA, P-Pert is a cross-model universal adversarial perturbation designed to protect multiple facial images. As shown in the quantitative results Table~\ref{tab:mainresult} and the visualizations in Fig.~\ref{fig:result}, P-Pert significantly outperforms the baseline universal perturbation in both facial region disruption and cross-model balance. This advantage stems from its use of a dynamic weighting strategy and an identity disruption module for optimization. Compared to the proposed generative perturbation, P-Pert achieves cross-image universality at the cost of a certain degree of reduced defense performance. Defenders can balance performance and computational efficiency by selecting either $\mathcal{G}$ or P-Pert for image protection.
    
\subsection{Ablation Study}

\subsubsection{Identity Destruction Module}
\begin{figure}[]
          \centering
          \includegraphics[width=\linewidth]{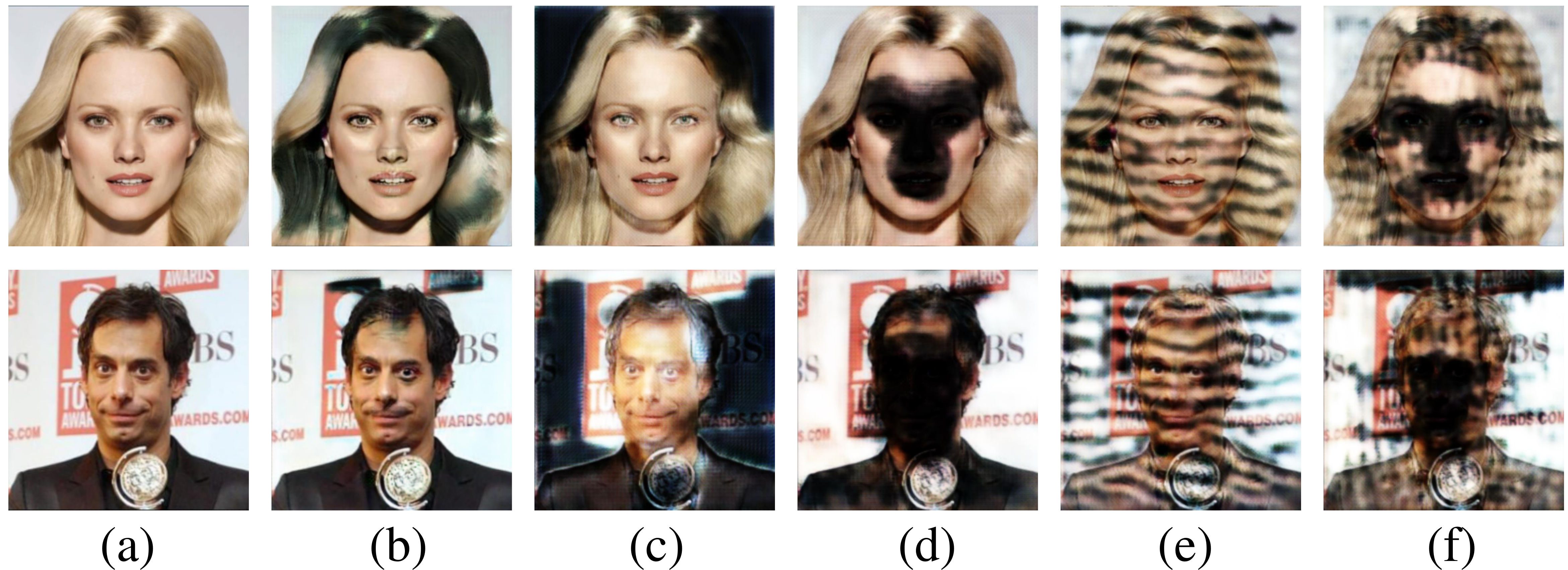}
          \caption{Visual examples of ablation study of Identity Destruction Module. Among them (a) is a natural image, (b) is a fake image, and (c)-(f) correspond to conditions \#1, \#3, \#4, and \#5 in Table \ref{tab:module}, respectively.}
          \label{fig:module}
    \end{figure}

The Identity Destruction Module aims to destroy the identity semantics of a face so that it cannot be correctly recognized. We delve into the impact of the three designed losses on the destruction effect. Table~\ref{tab:module} and Fig.~\ref{fig:module} present the quantitative and visual ablation results, respectively. Specifically, the three sub-modules focus on different issues. The mask-constrained loss uses two facial masks as strong constraints for the attack, thus providing a huge improvement in significantly distorting facial regions. Identity loss is a feature-level constraint that perturbs the key areas of identity recognition from a global perspective of the image. This design is important in destroying machine identification and will be introduced in detail in Section~\ref{Misleading Facial Recognition Systems}. As shown in Fig.~\ref{fig:module}, the mask-constrained loss concentrates the distortion on the face region of the image, while the identity consistency loss destroys the global texture. The feature confusion loss brings an overall gain, which benefits from the similarity in feature extraction of the facial manipulation model. It is worth noting that the three types of losses reinforce each other to some extent.
\begin{table}[]
    \centering
    \renewcommand{\arraystretch}{1.0}
    \caption{Ablation results for component modules. The best result in each column is marked in \textbf{bold}.}
    \label{tab:module}
    \begin{tabular}{cl|ccc}
        \toprule
        &Settings &$L_{2}^{face}$ $\uparrow$ &ID sim.$\downarrow$ &DSR$\uparrow$ \\
        \midrule
        \#1 &w/o all  &0.172 &0.509 &0.358\\
        \midrule
        \#2 &w/o FCL  &0.396 &0.050 &0.973\\
        
        \#3 &w/o ICL &\textbf{0.431} &0.148 &0.937\\
        
        \#4 &w/o MCL  &0.189 &0.039 &0.866\\ 
        \midrule
        \#5 &w/ all  &0.425 &\textbf{0.031} &\textbf{0.999}\\
        \bottomrule
    \end{tabular}
    
    \begin{tabular}{@{}p{10cm}}\\
        $\qquad \quad$ FCL means the \textbf{F}eature \textbf{C}onfusion \textbf{L}oss.\\
        $\qquad \quad$ ICL means the \textbf{I}dentity \textbf{C}onsistency \textbf{L}oss.\\
        $\qquad \quad$ MCL means the \textbf{M}ask-\textbf{C}onstrained \textbf{L}oss.
  \end{tabular}
\end{table}
     
\subsubsection{Dynamic Weighting Strategy}
The dynamic weighting strategy focuses on balancing the attack losses for different facial manipulations. We selected equivalent weight, prior weight, hard model mining (HMM)~\cite{guan2022defending}, and DTP~\cite{guo2018dynamic} as the baseline of the weight setting methods. The experimental results are reported in Table \ref{tab:optim}. The equivalent weight setting will cause the generated perturbations to overfit on the most vulnerable model architecture (e.g., StarGAN and FPGAN). Although HMM balances each model to a certain extent, it ignores the difference in model gradients and thus causes the degradation of average performance. Furthermore, it performs poorly in defending against HiSD, achieving only a 68.1\% DSR. Separate prior weight settings or KPI are unstable and difficult to set, so we cleverly blend the two in Strategy II and get stable training. The benefits of this are two-fold: 1) it reduces the difficulty of a prior setting, and only needs to determine a series of orders of magnitude to allow automatic optimization of parameters; 2) it makes the dynamic weight more stable. 
\begin{table}[]
    \renewcommand{\arraystretch}{1.2}
    \centering
    \caption{Comparison of defense success rates under different optimization strategies. The best result in each column is marked in \textbf{bold}, while the second best result is marked with an \underline{underline}.}
    \label{tab:optim}
    \resizebox{1.0\columnwidth}{!}{
    \begin{tabular}{l|ccccc|c}
        \toprule
        Optimizations &StarGAN &AGGAN &FPGAN &RelGAN &HiSD &Average \\
        \midrule
         Equivalent weight  &\textbf{1.000} &0.458 &\textbf{1.000} &0.951 &0.420 &0.766\\
         Prior weight  &0.967 &\underline{0.986} &\textbf{1.000} &0.913 &0.875 &\underline{0.948}\\
         HMM      &0.990 &0.982 &\underline{0.985} &0.986 &0.681 &0.925\\
         DTP      &\textbf{1.000} &0.894 &\textbf{1.000} &0.802 &0.885 &0.916\\
        \midrule
         MGDA-based (S-I) &\textbf{1.000} &\textbf{1.000} &\textbf{1.000} &\underline{0.998} &\textbf{1.000} &\textbf{0.999}\\
         KPI-based (S-II) &\textbf{1.000} &\textbf{1.000} &\textbf{1.000} &\textbf{1.000} &\underline{0.998} &\textbf{0.999}\\
         
        \bottomrule
    \end{tabular}}
\end{table}

\textcolor{mycolor}{The MGDA-based strategy achieves excellent performance; however, it introduces additional computational overhead, making training more expensive. To quantify this overhead, we report the average training time per iteration, as well as the GFLOPs consumed during training per input sample, in Table~\ref{tab:computational_overhead_optim}. Compared with the baseline, MGDA increases the training time by approximately 30\% due to the need for extra backward propagation through the perturbation generator to compute task-specific gradients. This results in an additional cost of 430.25 GFLOPs over the baseline. Although the theoretical GFLOPs grow significantly, the training time grows more modestly. This is because the additional cost mainly comes from repeated backward passes through a lightweight module, which modern GPUs handle efficiently, leading to a smaller impact on actual runtime. In contrast, the KPI-based strategy introduces minimal overhead, as it estimates the weights via a lightweight closed-form computation that avoids the gradient-based optimization.}

\begin{table}[]
    \renewcommand{\arraystretch}{1.0}
    \centering
    \caption{Quantitative comparison of average training time per iteration (batch size = 4) and GFLOPs consumed per input sample during training under different strategies.}
    \label{tab:computational_overhead_optim}
    \begin{tabular}{l|cc}
        \toprule
        \textbf{Optimizations} & \textbf{Training Time (s/iter) $\downarrow$} 
        & \textbf{GFLOPs/sample $\downarrow$} \\
        \midrule
         Equivalent weight  &1.24  &1670.22 \\
         \midrule
         MGDA-based (S-I) &1.50  &2100.47 \\
         KPI-based (S-II) &1.24  &1670.22 \\
        \bottomrule
      \end{tabular}
\end{table}

\subsubsection{Gradient Prior Perturbation}
\begin{figure}[]
          \centering
          \setcounter{subfigure}{0}
          \subfigure [Changes of training loss] {\includegraphics[width=.24\textwidth]{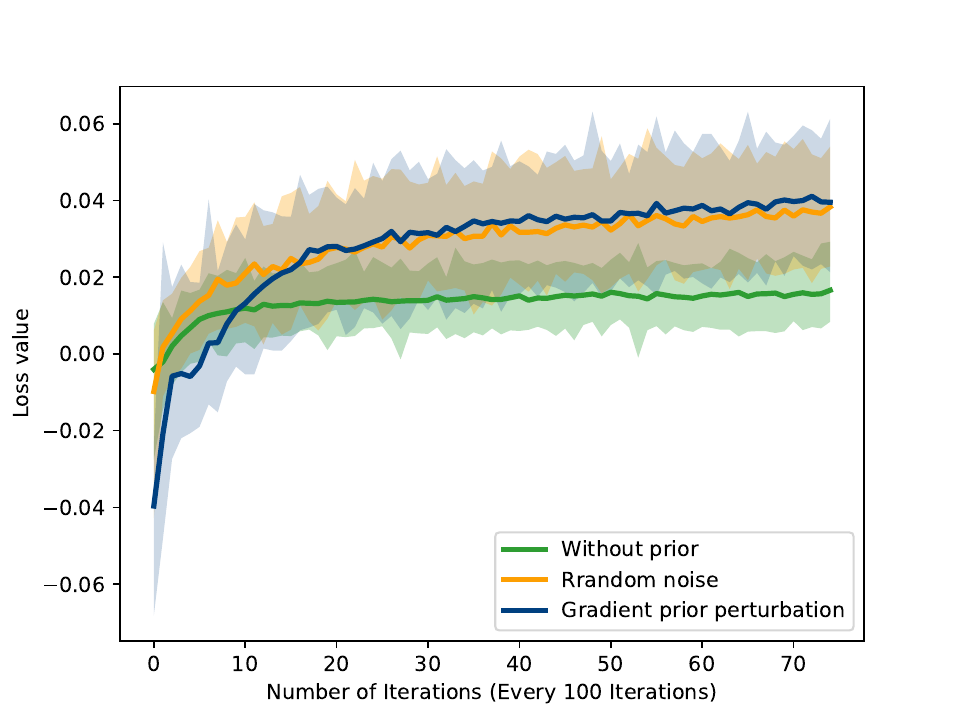}}
    	\subfigure [Defense results at different scales] {\includegraphics[width=.24\textwidth]{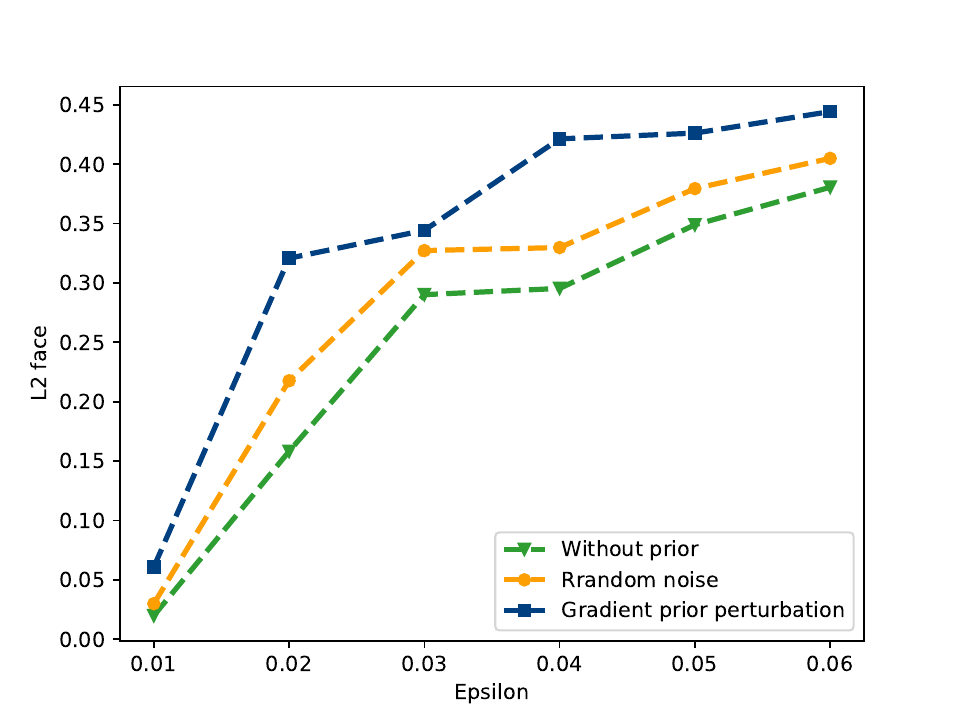}}
          \caption{The changes of training loss and defense performance at different scales with a gradient prior perturbation, with random noise, and without any prior knowledge.}
          \label{fig:ifpri}
    \end{figure}

Gradient prior perturbation aims to provide the generator with noise-like prior knowledge, thus accelerating its convergence. For comparison, the variation of training loss and defense performance at different scales with the gradient prior perturbation, with a prior random noise, and without prior knowledge is shown in Fig.~\ref{fig:ifpri}. Both gradient prior perturbation and random noise promote the convergence of the generator, which is due to the introduction of a global noise structure~\cite{aneja2022tafim}. In terms of generator performance, methods based on gradient prior perturbation at different scales have shown the most significant defense effect, with an average improvement of $31.2\%$ compared to random noise methods. We believe that the reason behind this is that the gradient prior perturbation involves rich adversarial structural information.

\subsubsection{Architecture of the Perturbation Generator}
We explore the impact of different generator architectures on performance. Three mainstream architectures, including Unet~\cite{ronneberger2015u}, Resnet~\cite{he2016deep}, and Transformer~\cite{vaswani2017attention}, are selected as the generators of the proposed ID-Guard. Fig.~\ref{tab:ach} reports the defense performance of the generators for these three architectures. Compared with Unet, Resnet, and Transformer architectures have achieved significant advantages. As shown in Table~\ref{tab:ach}, Transformer achieved optimal performance at the expense of model parameter size, while Resnet achieved very close performance with less than $5\%$ of its parameter size. We propose to use Resnet as the architecture for the generator of the proposed ID-Guard, and the intuition behind this is that the generated perturbation can be regarded as a residual of the image.
\begin{figure}[]
    \centering
            \setcounter{subfigure}{0}
    	\subfigure [$L_{2}^{face}$] {\includegraphics[width=.24\textwidth]{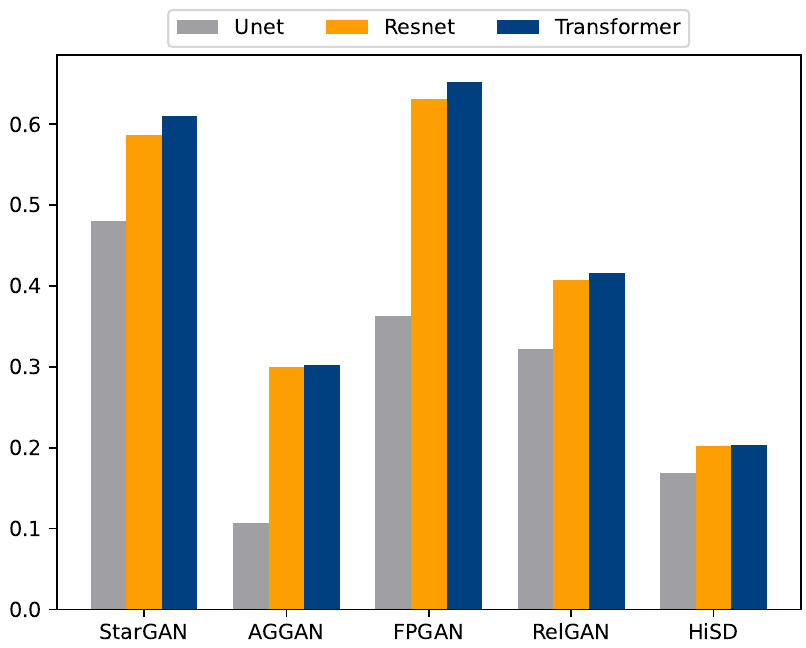}}
    	\subfigure [Defense success rate] {\includegraphics[width=.24\textwidth]{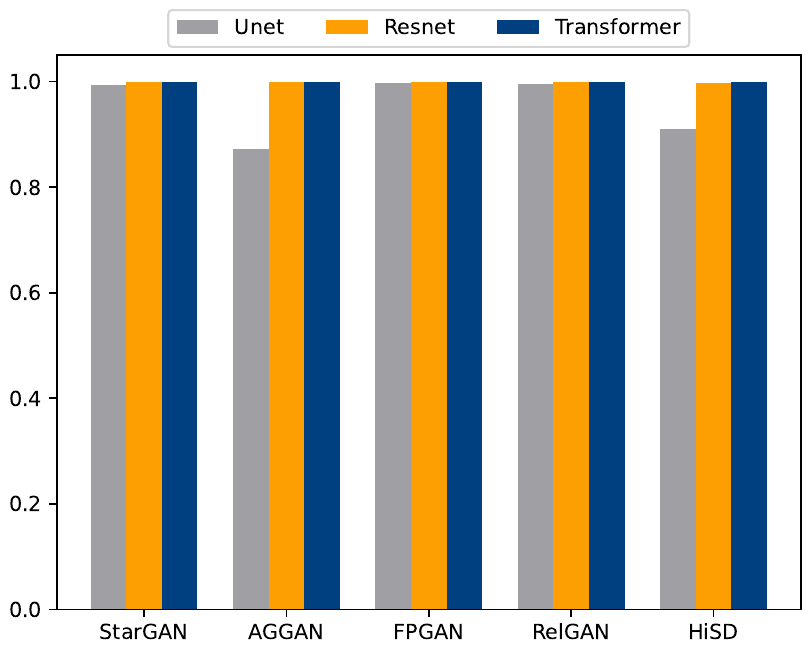}}
    	\caption{Performance comparison of perturbation generators based on different architectures.}
    	\label{fig:ach}
\end{figure}

\begin{table}[]
    \renewcommand{\arraystretch}{1.2}
		\centering
		\caption{Comparison of the number of model parameters for generators based on different architectures.} \label{tab:ach}	
			\begin{tabular}{lc}
                \toprule
                \textbf{Generator Architecture} & \textbf{\#Params}  \\
			\midrule
				Unet-based Generator              &54.4M\\
				Resnet-based Generator            &7.9M\\
				Transformer-based Generator       &179.3M\\
			\bottomrule
    \end{tabular}
\end{table}

\subsection{Other Evaluation}

\subsubsection{Misleading Facial Recognition Systems} \label{Misleading Facial Recognition Systems}
\begin{figure}[t]
      \centering
      \includegraphics[width=\linewidth]{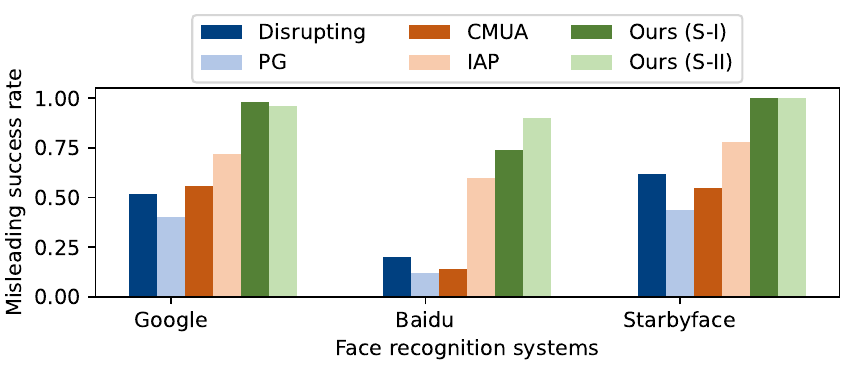}
      \caption{Quantitative comparison of the misdirection success rates of distorted images on three mainstream commercial facial recognition systems.}
      \label{fig:misleading}
\end{figure}

Some social applications recognize photos uploaded by users and then add corresponding tags and use them in content recommendation systems. This can exacerbate the spread of distorted faces. Therefore, the threat of stigmatization of distorted images comes not only from the human eye but also from commercial facial recognition systems. As shown in Fig.~\ref{fig:misleading}, we evaluate the misdirection success rates of the destroyed outputs of ID-Guard and competing algorithms on three mainstream facial recognition systems. As can be seen, our method reports optimal results, achieving an over-$95\%$ misdirection success rate on Google~\footnote{https://www.google.com} and StarByFace~\footnote{https://starbyface.com}. Baidu~\footnote{https://www.baidu.com} has the most robust recognition system, with CMUA~\cite{huang2022cmua} and PG~\cite{huang2021initiative} can hardly fool it, but ID-Guard still causes it to recognize more than $75\%$ of images incorrectly.
The good performance of ID-Guard is due to the identity consistency loss introduced to destroy the identity recognition baseline model, which is widely used in commercial facial recognition systems.

\begin{table}[]
    \renewcommand{\arraystretch}{1.2}
    \centering
    \caption{Quantitative results of the image inpainting on distorted images. The best result is marked in \textbf{bold}, while the second best result is marked with an \underline{underline}. }\label{tab:inpaintings}
    
    \resizebox{1.0\columnwidth}{!}{
        
        \begin{tabular}{c|c|ccccc}
        \toprule
            facial inpainting
            &Methods
            &StarGAN &AGGAN &FPGAN &RelGAN &HiSD \\
        \midrule
            
            \multirow{6}{*}{LBP \cite{wu2021deep}} 
            &Disrupting\cite{ruiz2020disrupting}           
            &\textbf{0.664} &0.114 &0.156 &0.063 &0.053 \\
            &PG\cite{huang2021initiative}          
            &0.095 &0.070 &0.092 &0.055 &0.059 \\
            &CMUA\cite{huang2022cmua}            
            &0.309 &0.089 &0.093 &0.162 &0.072 \\
            &IAP\cite{zhu2023information}             
            &0.303 &0.094 &0.266 &0.159 &\underline{0.080} \\
            \cmidrule(lr){2-7}
            &Ours (S-I)       
            &0.259 &\textbf{0.141} &\underline{0.335} &\underline{0.197} &\textbf{0.084}  \\
            &Ours (S-II)       
            &\underline{0.389} &\underline{0.137} &\textbf{0.344} &\textbf{0.218} &0.076 \\
            \midrule

            \multirow{6}{*}{GS-SSA \cite{sheng2024deep}} 
            &Disrupting\cite{ruiz2020disrupting}           
            &\textbf{0.772} &0.081 &0.106 &0.028 &0.017 \\
            &PG\cite{huang2021initiative}          
            &0.093 &0.033 &0.054 &0.019 &0.024 \\
            &CMUA\cite{huang2022cmua}            
            &0.437 &0.053 &0.064 &\underline{0.148} &0.044 \\
            &IAP\cite{zhu2023information}             
            &0.318 &0.065 &0.285 &0.134 &0.048 \\
            \cmidrule(lr){2-7}
            &Ours (S-I)       
            &0.296 &\underline{0.108} &\underline{0.328} &0.143 &\textbf{0.059}  \\
            &Ours (S-II)      
            &\underline{0.481} &\textbf{0.113} &\textbf{0.372} &\textbf{0.150} &\underline{0.058} \\
            \bottomrule
    \end{tabular}}
\end{table}

\begin{figure}[]
      \centering
      \includegraphics[width=\linewidth]{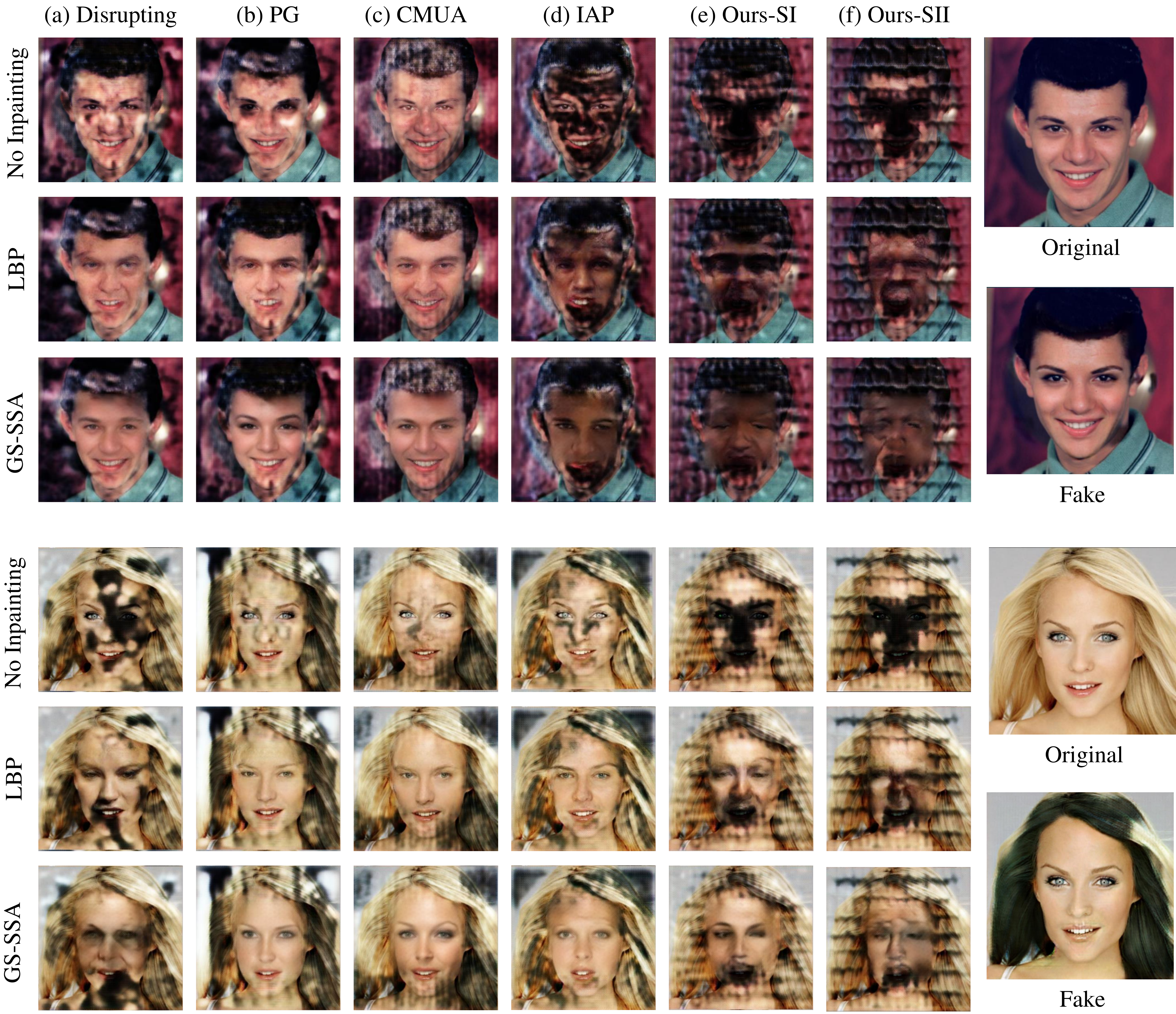}
      \caption{Visual examples of the image inpainting on distorted images. The original and fake images are provided on the right side for reference.}
      \label{fig:inpaintings}
\end{figure}

\subsubsection{Resisting facial inpainting}
Another challenge comes from the image inpainting system. A well-trained facial inpainting model can recover distorted facial images, rendering defenses ineffective. We evaluate the performance of distorted images against two baseline facial inpainting models, namely LBP~\cite{wu2021deep} and GS-SSA~\cite{sheng2024deep}. The quantitative results of $L_{2}^{face}$ distance are reported in Table~\ref{tab:inpaintings}, and the visualization results are shown in Fig.~\ref{fig:inpaintings}. Although the difference between the repaired distorted image and the forged result is greatly reduced, the proposed method still exhibits optimal defense performance. This is because these facial inpainting systems rely heavily on undistorted regions of the image. However, we achieve a greater degree of destruction of the entire image texture due to the introduction of feature loss.

\begin{table}
    \centering
    \caption{Summary of manipulation models used in training and evaluation.}
    \resizebox{\linewidth}{!}{%
    \begin{tabular}{llcccl}
    \toprule
    \textbf{Type} & \textbf{Models} & \textbf{Train} & \textbf{Struct.} & \textbf{Param.} & \textbf{Seen} \\
    \midrule
    White-box & \makecell[l]{StarGAN, AGGAN, FPGAN,\\RelGAN, HiSD} & \checkmark & \checkmark & \checkmark & Seen \\
    Gray-box  & StarGAN$^{\dagger}$, RelGAN$^{\dagger}$ & -- & \checkmark & -- & Unseen \\
    Black-box & GANimation & -- & -- & -- & Unseen \\
    \bottomrule
    \end{tabular}%
    }
    \\[3pt]
    $^{\dagger}$ The facial manipulation model in the gray-box setting.
    \label{tab:model_summary}
\end{table}

\subsubsection{Evaluation against Unseen Manipulation Models}
\textcolor{mycolor}{The goal of this section is to assess the transferability of ID-Guard to manipulation models that are not accessible during training. In other words, we aim to evaluate whether perturbations optimized on the white-box models can still provide effective defense when transferred to unseen models with different accessibility levels. To clarify which manipulation models are accessible during training and which are held out for evaluation, Table~\ref{tab:model_summary} summarizes the manipulation models used in this experiment. Models marked as ``Seen'' denote those included in the target model set $\mathcal{S}_{\mathcal{M}}$, which are accessible and used to train the perturbation generator; Models marked as ``Unseen'' are excluded from $\mathcal{S}_{\mathcal{M}}$ and held out for exclusive use in evaluating transferability. Specifically, in the \emph{gray-box} setting, the evaluated manipulation models share the same architecture and training paradigm as some white-box models in $\mathcal{S}_{\mathcal{M}}$, but have different internal parameters. We retrained StarGAN~\cite{choi2018stargan} and RelGAN~\cite{wu2019relgan} from scratch with different random seeds and data splits to serve as gray-box models (marked with $^{\dagger}$); In the \emph{black-box} setting, we adopt GANimation~\cite{pumarola2020ganimation}, a widely used facial reenactment model excluded from the training model set, for evaluation in this scenario.}

\textcolor{mycolor}{
For gray-box scenarios, as shown in Table~\ref{tab:gray}, Disrupting~\cite{ruiz2020disrupting} derives adversarial perturbations on the accessible white-box versions of StarGAN and RelGAN, respectively. This model-specific perturbation generation enables it to maintain gray-box transferability to some extent.
ID-Guard still maintains the most powerful defense capabilities. This is mainly because the IDM-oriented feature loss in ID-Guard destroys identity-relevant features instead of overfitting to the shallow gradient directions of a single model, making the resulting perturbations more transferable to unseen models.
Additionally, the baseline methods perform better against the gray-box RelGAN$^{\dagger}$ than against its white-box version. The reason for this could be that their imbalanced training process caused the perturbation performance to be biased toward the most vulnerable white-box StarGAN, which has adversarial gradients that are closer to the gray-box RelGAN$^{\dagger}$ used in this experiment. 
}

\begin{table}[]
        \renewcommand{\arraystretch}{1.2}
		\centering
		\caption{Quantitative results in gray-box scenarios. The best result is marked in \textbf{bold}, while the second best result is marked with an \underline{underline}. }\label{tab:gray}
		\resizebox{1.0\columnwidth}{!}{
			
			\begin{tabular}{c|cccccc}
			\toprule
                \multirow{2}{*}{Methods} 
                &\multicolumn{3}{c}{StarGAN$^{\dagger}$}  
                &\multicolumn{3}{c}{RelGAN$^{\dagger}$} \\
				\cmidrule(l){2-4} \cmidrule(l){5-7} &$L_{2}^{face}$ $\uparrow$ &ID sim.$\downarrow$ &DSR$\uparrow$ &$L_{2}^{face}$$\uparrow$ &ID sim.$\downarrow$ &DSR$\uparrow$ \\
				\midrule
				Disrupting\cite{ruiz2020disrupting}           
				&\textbf{0.187} &0.517 &0.234 &\textbf{0.201} &\underline{0.161} &\underline{0.832}\\
				PG\cite{huang2021initiative}          
				&0.094 &0.630 &0.068 &0.044 &0.619 &0.008\\
				CMUA\cite{huang2022cmua}            
				&0.090 &0.798 &0.035 &0.041 &0.719 &0.008\\
				IAP\cite{zhu2023information}             
				&0.088 &0.573 &0.117 &0.096 &0.372 &0.421\\
				\cmidrule(lr){1-7} 
				Ours (S-I)      
				&\underline{0.137} &\underline{0.305} &\textbf{0.620} &0.168 &0.207 &0.724\\
				Ours (S-II)       
				&0.133 &\textbf{0.329} &\underline{0.581} &\underline{0.176} &\textbf{0.100} &\textbf{0.906}\\
				\bottomrule
		\end{tabular}}
       $^{\dagger}$ The facial manipulation model in the gray-box setting.
\end{table}

\begin{figure}
      \centering
      \includegraphics[width=0.7\linewidth]{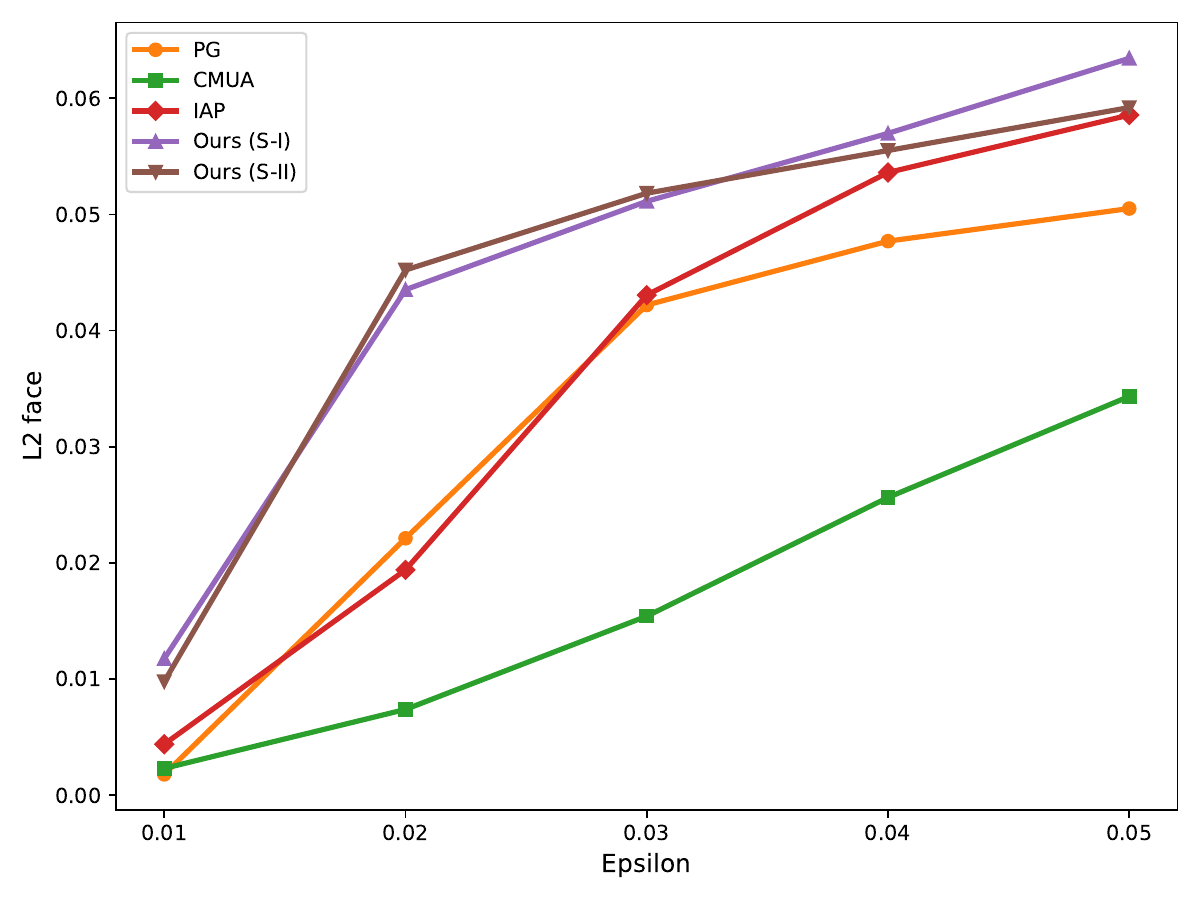}
      \caption{The performance of defense methods against GANimation under different perturbation thresholds.}
      \label{fig:black_box}
\end{figure}

\textcolor{mycolor}{
For black-box evaluation, we report in Fig.~\ref{fig:black_box} the performance of ID-Guard and other transferable defense methods against GANimation under different perturbation thresholds. The proposed ID-Guard consistently achieves the best (largest) $L_{2}^{face}$ distances compared to the baselines; the advantage is particularly pronounced at small thresholds ($\varepsilon \leq 0.03$). We first attribute this to the designed feature confusion loss, which disrupts model feature extraction with cross-model consistency~\cite{tang2023feature}. Second, the dynamic weight training strategy and diverse $\mathcal{S}_{\mathcal{M}}$ reduce overfitting to any single white-box model and favor perturbation patterns that break common identity representations across multiple models. In addition, ID-Guard’s emphasis on fine-grained feature destruction yields stronger transferability even when the perturbation magnitude is constrained. Nonetheless, the overall performance against the black-box model exhibits a noticeable drop, suggesting a potential trade-off between strong white-box cross-model ability and black-box transferability.
}

\begin{figure}[]
    \centering
    \setcounter{subfigure}{0}
    \subfigure [JPEG compression] {\includegraphics[width=.24\textwidth]{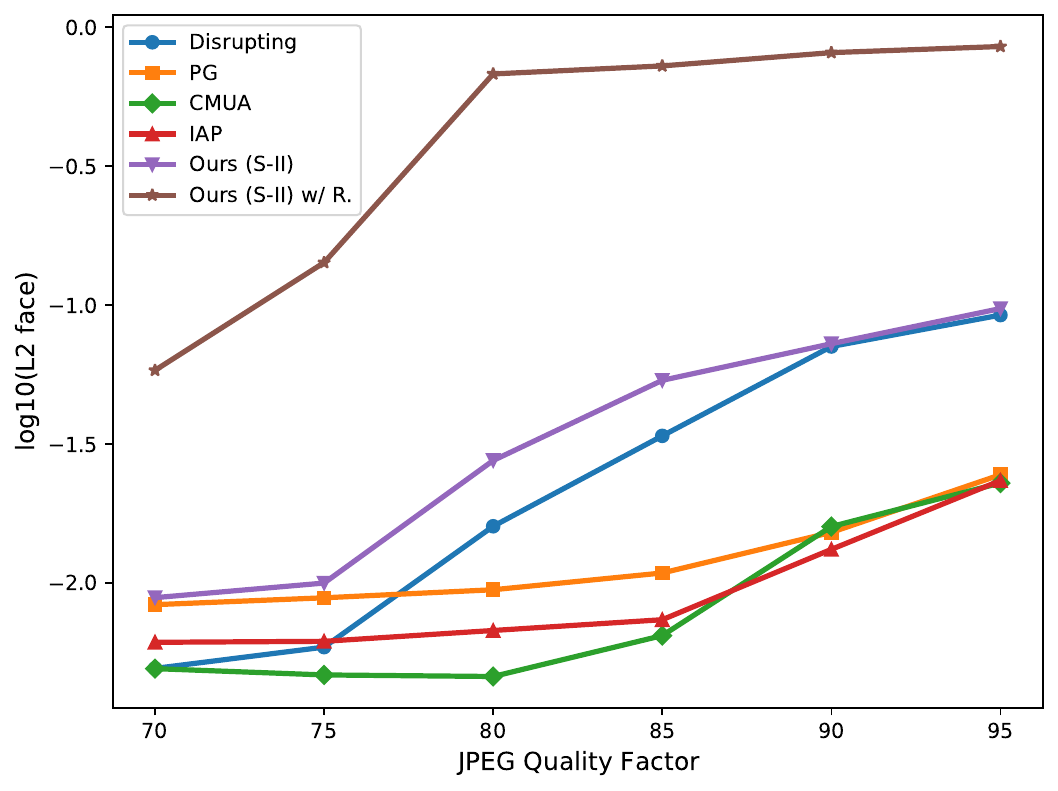}}
    	\subfigure [Gaussian Blur] {\includegraphics[width=.24\textwidth]{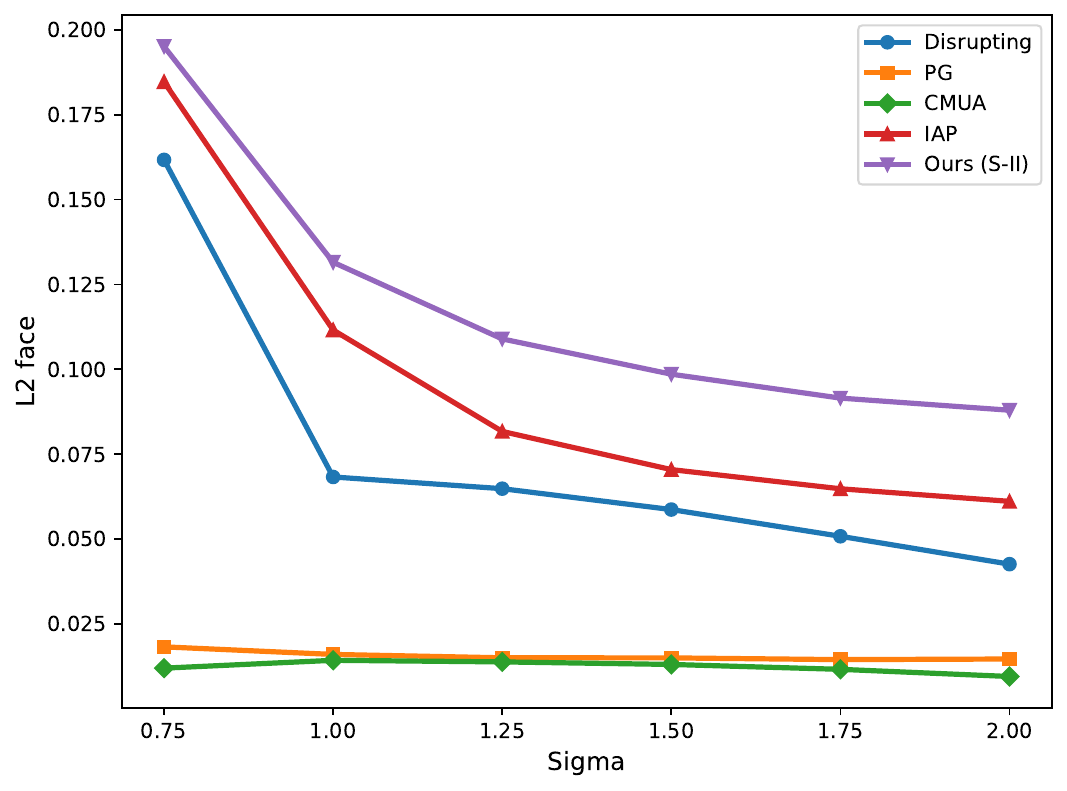}}
    	\caption{The performance of the defense algorithm against StarGAN under different intensities of lossy operations. In the JPEG evaluation, the performance of ID-Guard integrated with the compression-resistant strategy has a significant advantage. Therefore, to present the results, we take the logarithm of 10 for $L_{2}^{face}$.}
    	\label{fig:com_blur}
\end{figure}

\begin{figure}[]
      \centering
      \includegraphics[width=\linewidth]{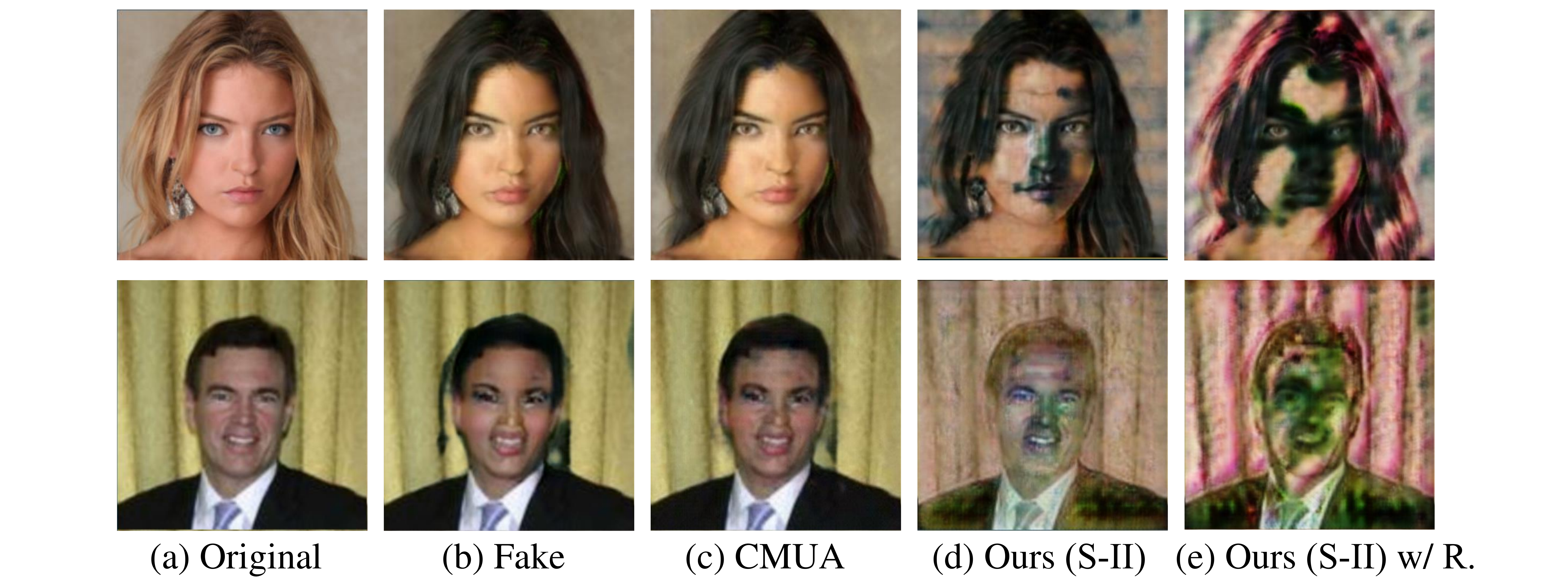}
      \caption{Visualization of adversarial examples against the StarGAN model under JPEG compression with QF=85.}
      \label{fig:osns}
\end{figure}

\begin{table}[]
    \renewcommand{\arraystretch}{1.0}
    \centering
    \caption{Quantitative impact of the compression-resistant strategy on average training time per iteration (batch size = 4) and GFLOPs consumed per input sample during training.}

    \label{tab:computational_overhead_comp}
    \begin{tabular}{c|cc}
        \toprule
         & \textbf{Training Time (ms/iter) $\downarrow$} 
        & \textbf{GFLOPs/sample $\downarrow$} \\
        \midrule
         w/o R.  &1.24  &1670.22 \\
         w/ R.   &1.25  &1751.39 \\
        \bottomrule
    \end{tabular}
\end{table}

\begin{table*}
    \centering
    \caption{Quantitative results of the adversarial robustness of the original facial manipulation model and its adversarial training version against different attacks.}
        \label{tab:advtrain}
        \begin{tabular}{c|cccccccccc}
            \toprule
            \multirow{2}{*}{Model} 
            &\multicolumn{2}{c}{Gaussian noise ($\sigma=0.05$)}  
            &\multicolumn{2}{c}{FGSM} 
            &\multicolumn{2}{c}{I-FGSM}
            &\multicolumn{2}{c}{PGD}
            &\multicolumn{2}{c}{C\&W}
            \\
            \cmidrule(l){2-3} \cmidrule(l){4-5} \cmidrule(l){6-7} \cmidrule(l){8-9} \cmidrule(l){10-11} &$L_{2}^{face}$ $\downarrow$ &ID sim.$\uparrow$  &$L_{2}^{face}$$\downarrow$ &ID sim.$\uparrow$ &$L_{2}^{face}$$\downarrow$ &ID sim.$\uparrow$ &$L_{2}^{face}$$\downarrow$ &ID sim.$\uparrow$ &$L_{2}^{face}$$\downarrow$ &ID sim.$\uparrow$\\
            \midrule
            StarGAN &0.070 &0.261 &0.185 &0.184 &1.075 &0.069 &1.102 &0.004 &1.377 &0.009\\
            StarGAN-AT &0.001 &0.985 &0.011 &0.977 &0.148 &0.700 &0.141 &0.712 &0.149 &0.756\\
            \midrule
            RelGAN &0.003 &0.905 &0.140 &0.766 &0.584 &0.186 &0.615 &0.145 &0.983 &0.099\\
            RelGAN-AT &0.000 &0.992 &0.001 &0.989 &0.019 &0.783 &0.018 &0.791 &0.002 &0.988\\
            \bottomrule
    \end{tabular}
\end{table*} 

\subsubsection{Robustness under Lossy Operations}
In real scenarios, users often upload perturbed images to social applications to share their lives. However, various lossy operations on the transmission channel can destroy the effectiveness of the perturbation. In this section, we evaluate the robustness of ID-Guard and competing algorithms under JPEG compression and Gaussian blur. To further verify the integration capability of our framework and the robustness strategy, for JPEG, we incorporate the compression-resistant strategy from \cite{qu2024df} into our generator training. The results are reported as ``Ours (S-II) w/ R.". As shown in Fig.~\ref{fig:com_blur}, as the intensity of the lossy operation increases, the defense performance of each method is gradually weakened. Our method demonstrates significant robustness at different scales. The underlying reason for this could be the introduction of the identity disruption module, which concentrates the effectiveness of the adversarial perturbations in specific areas, making them less susceptible to degradation from lossy operations. When the robustness training strategy was integrated, ID-Guard’s robustness improved significantly. A sample visualization is shown in Fig.~\ref{fig:osns}. This demonstrates the flexibility of the proposed framework and its ability to effectively integrate with advanced strategies from the research community.

\begin{figure}
      \centering
      \includegraphics[width=\linewidth]{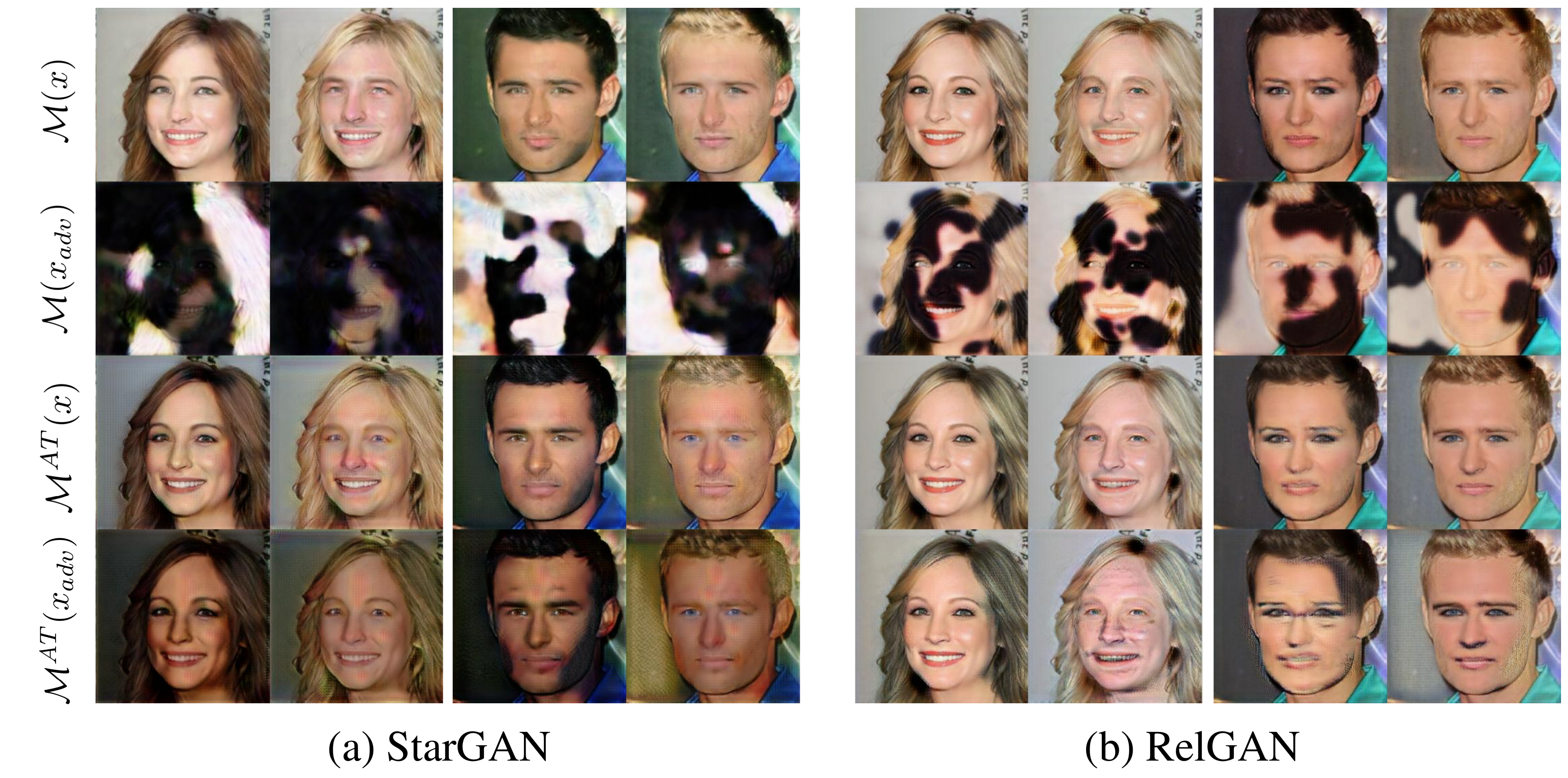}
      \caption{Visual example of the adversarial robustness of the original facial manipulation model and its adversarial training version against PGD\cite{madry2017towards}.}
      \label{fig:advtrain}
\end{figure}

\textcolor{mycolor}{We report in Table~\ref{tab:computational_overhead_comp} the impact of incorporating the compression-resistant strategy on the training time and computational overhead of the perturbation generator. The results show that this strategy introduces negligible change in per-iteration training time, and only increases the FLOPs by approximately 5\%. Therefore, we recommend adopting the compression-resistant strategy when deploying the proposed framework in online scenarios.}

\subsubsection{ID-Guard for Adversarial Training}
\label{ID-Guard for Adversarial Training}
In this section, we evaluate the effectiveness of using the ID-Guard framework for adversarial training of facial manipulation models. 
Specifically, we select StarGAN~\cite{choi2018stargan} and RelGAN~\cite{wu2019relgan} as example facial manipulation models and apply the adversarial training (AT) approach proposed in Section~\ref{AT} to obtain StarGAN-AT and RelGAN-AT. The adversarial training version of the facial manipulation model $\mathcal{M}$ is noted as $\mathcal{M}^{AT}$. The training framework follows their official open-source implementations~\footnote{https://github.com/yunjey/stargan, https://github.com/elvisyjlin/RelGAN-PyTorch}. 
To evaluate adversarial robustness, we test these models against Gaussian noise ($\sigma=0.05$), FGSM~\cite{goodfellow2014explaining}, I-FGSM~\cite{kurakin2018adversarial}, PGD~\cite{madry2017towards}, and C\&W~\cite{carlini2017towards} attacks. All attacks follow the white-box setup. As reported in Table~\ref{tab:advtrain}, non-adversarial Gaussian noise and the single-step FGSM disrupt $\mathcal{M}$ to some extent but have almost no impact on $\mathcal{M}^{AT}$. Furthermore, for the three stronger attack algorithms, I-FGSM, PGD, and C\&W, the model implemented adversarial training demonstrates significant robustness. Especially for RelGAN-AT, the $L_2^{face}$ consistently remains within the threshold of $0.05$. Additionally, the quantitative identity similarity results and the visualized examples in Fig.~\ref{fig:advtrain} indicate that these adversarial attacks mainly distort the background regions of $\mathcal{M}$'s manipulated outputs while leaving the facial regions unaffected. This validates that the proposed ID-Guard framework can serve as a plug-and-play adversarial attack module within adversarial training, significantly enhancing the robustness of facial manipulation models.

\begin{figure}
    \centering
            \setcounter{subfigure}{0}
            \includegraphics[width=.4\textwidth]{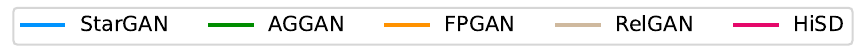}
    	\subfigure [Strategy I] {\includegraphics[width=.24\textwidth]{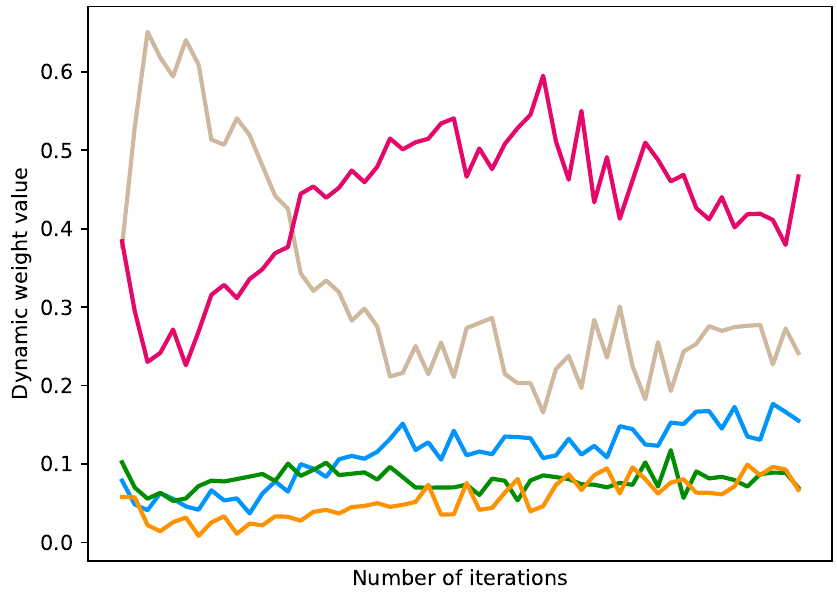}}
    	\subfigure [Strategy II] {\includegraphics[width=.24\textwidth]{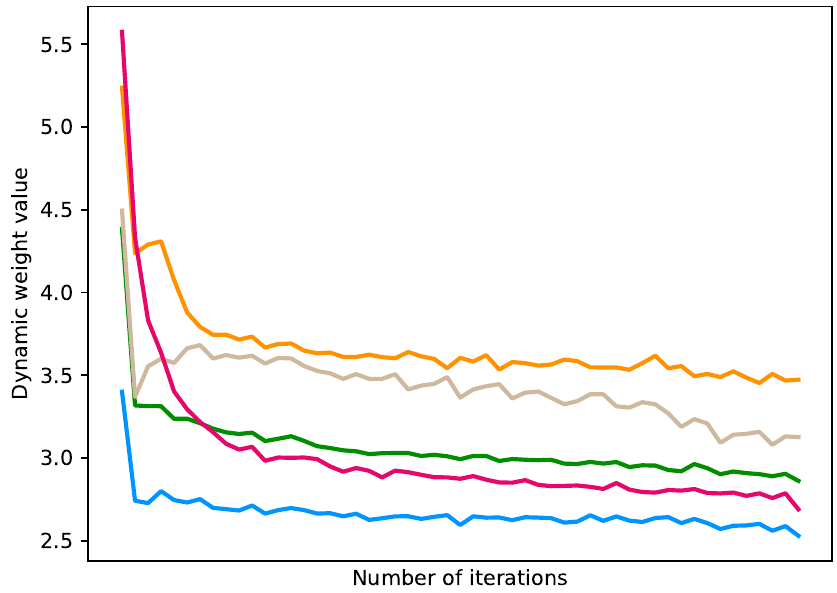}}
    	\caption{The changing trend of the weight of attack loss with the number of iterations for different facial manipulations when adopting the dynamic weighting strategy.}
     \label{fig:dws}	
\end{figure}

\subsection{Further Discussion} \label{Further Discussion}

\subsubsection{How Weights Dynamically Change?}
In the proposed ID-Guard framework, the dynamic weighting strategy is very important, which directly affects the training process and the balance of attack losses for different facial manipulations. Here, to explore its mechanism in depth, we record the dynamic changes of the weight set $\mathcal{S}_{\lambda} = \left\{ \lambda_{1},\lambda_{2},\dots,\lambda_{N} \right\}$ of attack losses in an epoch training, as shown in Fig.~\ref{fig:dws}. For strategy I, we leverage the MGDA algorithm to optimize the set of weights automatically in each iteration. It can be found that facial manipulation models with strong robustness, such as HiSD~\cite{li2021image} and RelGAN~\cite{wu2019relgan}, tend to be assigned larger weights to strengthen the attack against them. In addition, due to the lack of prior knowledge guidance, the weight change of strategy I is more affected by the current state of the generator, and thus fluctuates more significantly. For strategy II, in the initial stage, there is a large difference between the weights. With the constraints of the prior weights, the allocation of each dynamic weight stabilizes to obtain a balanced performance.

\subsubsection{Can ID-Guard be adapted for proactive forensics?}
To demonstrate the scalability of the proposed ID-Guard framework, we adapt it to proactive forensics in the context of facial manipulation. We follow the setup in \cite{zhu2023information}: Given a natural image $x$ and a binary message $m \in \{0,1\}^L$ of length $L$, a message embedding encoder $E_m$ maps the image and a corresponding message to a watermarked image. After the watermarked image undergoes manipulation, a message extractor decoder $D_m$ is used to recover $m$ from the forged image. The training objective is to minimize $\sum_{k=1}^N L^{wm}_k=MSE(m, D_m(\mathcal{M}_k(E_m(x,m))))$, while an additional discriminator is used to ensure the visual quality of the watermarked image. FaceTagger~\cite{wang2021faketagger} and information-based IAP~\cite{zhu2023information} are selected as baselines. In our method, the proposed dynamic weighting strategy is applied to the weight $L^{wm}_k$ for different
$\mathcal{M}_k$ during training. The Bit Error Rate (BER) and the visual quality of the watermarked image are quantitatively reported in Table~\ref{tab:forensic}. FaceTagger is trained with StarGAN~\cite{choi2018stargan} as the target model, leading to overfitting and consequently poor overall performance. Compared to IAP, our method demonstrates more balanced forensic performance across different facial manipulation models, particularly against HiSD~\cite{li2021image}, which exhibits significant structural differences. This result highlights the strong scalability and transferability of the proposed framework, making it a plug-and-play tool adaptable to various cross-model tasks in the community.

\begin{table}[]
        \renewcommand{\arraystretch}{1.2}
		\centering
		\caption{Quantitative results of proactive forensics. The best result is marked in \textbf{bold}.}\label{tab:forensic}
		\resizebox{1.0\columnwidth}{!}{
		\begin{tabular}{c|ccccc|c}
			\toprule
                \multirow{2}{*}{Methods} 
                &\multicolumn{5}{c|}{BER$\downarrow$}  
                &\multirow{2}{*}{SSIM$\uparrow$} \\
			\cmidrule(lr){2-6} &StarGAN  &AGGAN & FPGAN &RelGAN &HiSD \\
			\midrule
                FaceTagger\cite{wang2021faketagger} &\textbf{0.000} &0.501 &0.009 & 0.069 &0.512 &\textbf{0.995}\\
                IAP\cite{zhu2023information} &0.034 &0.144 &\textbf{0.000} &0.001 &\textbf{0.000} &0.983 \\
                Ours &\textbf{0.000} &\textbf{0.000} &\textbf{0.000} &\textbf{0.000} &\textbf{0.000} &\textbf{0.995} \\ 
			\bottomrule
		\end{tabular}}
\end{table}

\subsubsection{What does the generator learn?}
\begin{figure}[]
          \centering
          \includegraphics[width=\linewidth]{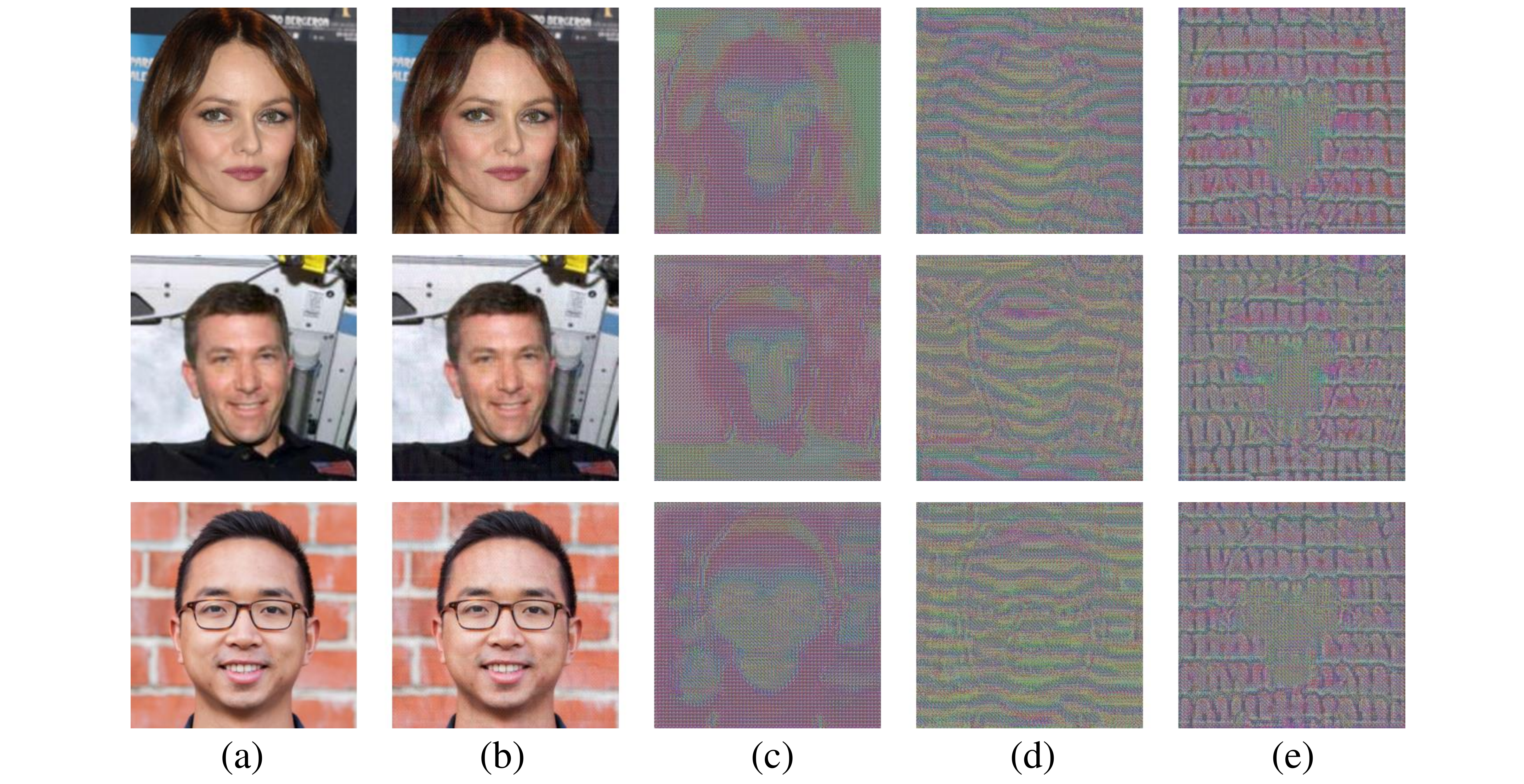}
          \caption{Visual examples of the generated perturbation. Among them, (a) is the original image, (b) is the adversarial image, (c) is the perturbation generated by the generator trained only with mask-constrained loss, (d) is the perturbation generated by the generator trained only with identity consistency loss, and (e) is the perturbation generated by the standard generator.}
          \label{fig:pertub}
    \end{figure}

The adversarial perturbations generated by the generator trained under different loss constraints are shown in Fig.~\ref{fig:pertub}. The mask-constrained loss concentrates rich adversarial information on the facial area of the image, thus completely distorting the output face. The generator trained using only the identity loss learns to disrupt images at the texture level. Combined with the results in Fig.\ref{fig:module}, this destruction changes the key feature semantics and visual attributes of the face. Therefore, the standard generator learns to generate adversarial perturbations that cause maximum damage to facial regions and alter the identifiable texture features of the image.

\section{Conclusion}   \label{Section V}
In this work, we proposed a universal framework for combating facial manipulation, named ID-Guard. To prevent face stigmatization problems caused by unconstrained image distortion, we propose an Identity Destruction Module to eliminate identity semantics. Furthermore, to improve the cross-model performance of generating perturbations, we regard attacking different models as a multi-task learning problem and introduce a dynamic parameter strategy. The proposed method not only effectively resists multiple facial manipulations but also significantly disrupts face identification. In addition, the experiment also demonstrated the possibility of ID-Guard in circumventing commercial facial recognition systems and image inpaintings. We hope that ID-Guard, with its good integration capabilities and application flexibility, can provide the community with an effective solution against facial manipulation.
                                                        
% \balance
\bibliographystyle{IEEEtran}
\bibliography{ref}

\end{document}